\documentclass[journal]{IEEEtranTIE}
\usepackage{graphicx}
\usepackage{cite}
\usepackage{picinpar}
\usepackage{amsmath}
\usepackage{url}
\usepackage{flushend}
\usepackage[latin1]{inputenc}
\usepackage{colortbl}
\usepackage{soul}
\usepackage{multirow}
\usepackage{pifont}
\usepackage{color}
\usepackage{alltt}
\usepackage[hidelinks]{hyperref}
\usepackage{enumerate}
\usepackage{siunitx}
\usepackage{breakurl}
\usepackage{epstopdf}
\usepackage{pbox}
\usepackage{amsmath,amsfonts}
\usepackage{graphicx}
\usepackage[caption=false,font=footnotesize]{subfig} % 不要与 caption/subcaption 同时使用

\usepackage{makecell} % 提供 \Xhline
\usepackage[table]{xcolor} 
\usepackage{xcolor}

\begin{document}
\title{	Diff-GNSS: Diffusion-based Pseudorange Error Estimation}

\author{
	\vskip 1em
	
	Jiaqi Zhu$^\dagger$,
	Shouyi Lu$^\dagger$,
    Ziyao Li,
    Guirong Zhuo*,
	  and Lu Xiong

	\thanks{
	
		Manuscript received Month xx, 2xxx; revised Month xx, xxxx; accepted Month x, xxxx.
		This work was supported by the National Natural Science Foundation of China under Grant 52325212, the National Key Research and Development Program of China under Grant 2022YFE0117100. (Corresponding author: Guirong Zhuo.)
        Jiaqi Zhu, Shouyi Lu, Ziyao Li, Guirong Zhuo, and Lu Xiong are with the School of Automotive Studies, Tongji University, Shanghai 201804, China (e-mails: 1911056@tongji.edu.cn; 2210803@tongji.edu.cn; 2231612@tongji.edu.cn; zhuoguirong@tongji.edu.cn; xiong\_lu@tongji.edu.cn).

	}
}

\maketitle
	
\begin{abstract}
Global Navigation Satellite Systems (GNSS) are vital for reliable urban positioning. However, multipath and non-line-of-sight reception often introduce large measurement errors that degrade accuracy. Learning-based methods for predicting and compensating pseudorange errors have gained traction, but their performance is limited by complex error distributions. 
% To address this challenge, we propose Diff-GNSS, a coarse-to-fine pseudorange error estimation framework that leverages a conditional diffusion model to capture such complex distributions. 
To address this challenge, we propose Diff-GNSS, a coarse-to-fine GNSS measurement (pseudorange) error estimation framework that leverages a conditional diffusion model to capture such complex distributions.
Firstly, a Mamba-based module performs coarse estimation to provide an initial prediction with appropriate scale and trend. Then, a conditional denoising diffusion layer refines the estimate, enabling fine-grained modeling of pseudorange errors. To suppress uncontrolled generative diversity and achieve controllable synthesis, three key features related to GNSS measurement quality are used as conditions to precisely guide the reverse denoising process. We further incorporate per-satellite uncertainty modeling within the diffusion stage to assess the reliability of the predicted errors. We have collected and publicly released a real-world dataset covering various scenes. Experiments on public and self-collected datasets show that Diff-GNSS consistently outperforms state-of-the-art baselines across multiple metrics. To the best of our knowledge, this is the first application of diffusion models to pseudorange error estimation. The proposed diffusion-based refinement module is plug-and-play and can be readily integrated into existing networks to markedly improve estimation accuracy.
\end{abstract}


\markboth{IEEE TRANSACTIONS ON INDUSTRIAL ELECTRONICS}%
{}

\definecolor{limegreen}{rgb}{0.2, 0.8, 0.2}
\definecolor{forestgreen}{rgb}{0.13, 0.55, 0.13}
\definecolor{greenhtml}{rgb}{0.0, 0.5, 0.0}

\section{Introduction}

\IEEEPARstart{A}{ccurate} positioning technologies are fundamental to modern transportation systems~\cite{ref01,ref41}, including autonomous vehicles~\cite{ref43,ref46} and unmanned aerial vehicles~\cite{ref42}. Due to its cost-effectiveness and global coverage, the Global Navigation Satellite System (GNSS) is extensively utilised for positioning~\cite{ref02}. However, GNSS measurements are susceptible to environmental effects, such as urban canyons~\cite{ref03} and wooded areas, that cause multipath and non-line-of-sight (NLOS) reception~\cite{ref04}. These phenomena can produce pseudorange errors of tens to hundreds of metres, severely constraining the reliability of GNSS positioning in urban areas~\cite{ref05,ref44}.

% To mitigate the adverse effects of large GNSS measurement errors on positioning, a variety of approaches have been investigated. Early studies have modelled multipath using signal quality indicators, such as the signal-to-noise ratio (SNR) and the carrier-to-noise density ratio (${C/N_{0}}$)~\cite{ref06,ref07}. Nevertheless, these methods tend to be ineffective in the presence of strong multipath. Another research line exploits environmental priors to identify and reject non-visible satellites, thereby filtering pseudorange measurements with large errors. This can be achieved by constructing 3D building models~\cite{ref08} or applying sky masks from fisheye cameras~\cite{ref09}, for instance. Such methods can improve positioning performance. However, these approaches rely heavily on the availability of 3D building models or additional sensors~\cite{ref10}.

Efforts to mitigate the adverse effects of large pseudorange errors on positioning have explored modeling multipath using signal quality indicators, such as the signal-to-noise ratio (SNR) and the carrier-to-noise density ratio (${C/N_{0}}$)~\cite{ref06,ref07}. However, these cues become unreliable under severe multipath. Environmental priors are also used to identify non-visible satellites and reject large-error pseudorange measurements, e.g., via 3D building models~\cite{ref08} or fisheye-camera sky masks~\cite{ref09}. While improving positioning, these methods depend on accurate 3D building models or additional sensors, limiting practicality~\cite{ref10}.

% In recent years, considerable interest has been generated in machine-learning-based methods for predicting GNSS measurement errors, whose complete and accurate physical modelling is challenging.~\cite{ref11} uses the gradient-boosted decision tree (GBDT) method to predict pseudorange errors based on signal strength, satellite elevation angle and pseudorange residuals. Deep learning based methods have also shown impressive performance due to their representation-learning capability.~\cite{ref12} uses fully connected neural networks (FCNNs) and long short-term memory (LSTM) networks to classify satellite visibility and predict pseudorange errors. Subsequently, promising improvements in positioning performance are achieved by using the network results to exclude NLOS measurements. Building on this work,~\cite{ref13} has incorporated self-attention into LSTM to enhance the utilisation of contextual information. While this markedly improves the classification of visibility, it does not improve the prediction of pseudorange error. This task remains challenging because pseudorange errors exhibit complex distributions, which are difficult for current networks to learn.

Recent work has applied machine learning to predict pseudorange errors.~\cite{ref11} uses the gradient-boosted decision tree (GBDT) method to predict pseudorange errors based on signal strength, satellite elevation angle and pseudorange residuals. Deep learning based methods further exploit representation learning: fully connected neural networks (FCNNs) and long short-term memory (LSTM) networks classify satellite visibility and predict pseudorange errors, and using their outputs to exclude NLOS improves positioning performance~\cite{ref12}. Building on this, self-attention is integrated into LSTM to better leverage context~\cite{ref13}, which improves visibility classification but not error regression. Accurate pseudorange error prediction remains challenging due to complex pseudorange error distributions.

\begin{figure}[!t]
\centering
% \vspace{-5pt}
\includegraphics[width=0.8\linewidth]{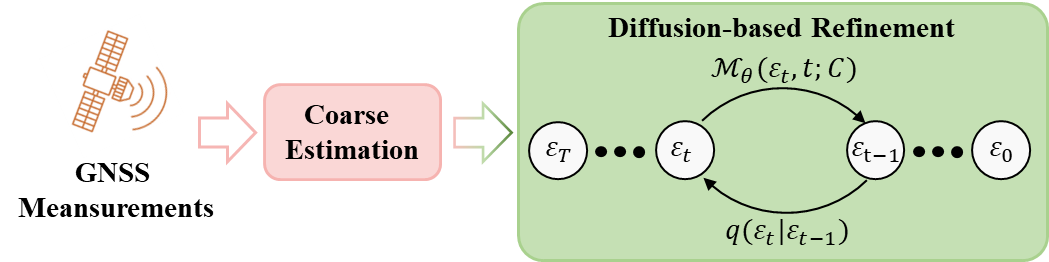}
\vspace{-8pt}
% \caption{The schematic of our coarse-to-fine GNSS measurement error estimation framework based on conditional diffusion. In the forward process, Gaussian noise is progressively added to the ground-truth pseudorange residual ${\varepsilon _{0}}$. A neural network ${\mathcal{M}_{\theta }  \left ( \varepsilon _{t},t;C  \right )  }$ is trained to denoise the noisy residual ${\varepsilon _{t}}$ at time step ${t}$ under the guidance of the conditioning signal ${C}$.}
\caption{The schematic of our coarse-to-fine pseudorange error estimation framework based on conditional diffusion.}
\vspace{-12pt}
\label{fig:simple_system}
\end{figure}

Recently, it has been found that diffusion models can learn complex distributions by gradually adding noise, which has achieved significant success in areas such as image synthesis~\cite{ref15}, object detection~\cite{ref16}, point cloud registration~\cite{ref17}, and scene flow estimation~\cite{ref18}. Conditioning signals can effectively guide the diffusion process for controllable generation. 
% In addition, diffusion models are also beneficial for escaping local optima by injecting random noise~\cite{ref19}.
This provides us with new ideas for our research. However, to the best of our knowledge, diffusion has not yet been successfully applied to the GNSS. Inspired by these developments, we propose Diff-GNSS, a method that redefines pseudorange error estimation as a conditional diffusion task, as shown in Fig.~\ref{fig:simple_system}. Our method follows a coarse-to-fine paradigm: a diffusion-based refinement module acts as a plug-and-play component that can be integrated into existing pseudorange error predictors.

However, diffusion models are a type of method with generative diversity, so applying them to the task of predicting deterministic pseudorange error is challenging. This is because we need to accurately calculate the error in the pseudorange measurements of each satellite. To address this challenge, we leverage GNSS spatio-temporal condition signals to guide controllable generation effectively. Specifically, we first obtain an initial pseudorange error via a Mamba-based coarse estimator, and then refine it by generating pseudorange error residuals with a conditional diffusion model. In the diffusion-based refinement stage, spatio-temporal features together with the coarse predictions serve as conditioning signals, allowing the network to learn a probabilistic mapping from the conditions to the residuals. In addition, it is essential to be able to aware the reliability of network learning results, a topic that has rarely been addressed in previous work. Inspired by the confidence modelling for scene flow in~\cite{ref18}, we introduce per-satellite uncertainty in order to evaluate the reliability of the pseudorange error estimates. Overall, the contributions of this paper are as follows:

\begin{itemize}
\item{To achieve reliable pseudorange error prediction, we introduce a novel coarse-to-fine framework that couples a Mamba-based coarse estimator with a diffusion-based refinement module. To the best of our knowledge, this is the first application of diffusion models to pseudorange error estimation. The designed diffusion-based refinement module has the advantage of being plug-and-play.}
\item{Within the diffusion-based refinement module, strong conditional guidance is achieved by extracting spatiotemporal features from raw GNSS measurements and embedding the coarse estimate.}
\item{We incorporate per-satellite uncertainty estimation to effectively guide the network in achieving reliable uncertainty aware.}
\item{We have collected and publicly released a real-world dataset\footnote{https://github.com/zhujq0220/Diff-GNSS} covering various scenes, which provides the community with a reproducible benchmark for fair evaluation and comparison.}
\item{Extensive evaluations on public and self-collected datasets demonstrate that our method consistently outperforms state-of-the-art (SOTA) baselines, validating the effectiveness of the framework. Furthermore, correcting pseudorange errors accurately significantly improves GNSS positioning accuracy.}
\end{itemize}

% The rest of the paper is organized as follows. Section~\ref{sec:related work} reviews related work. Section~\ref{sec:Diff-GNSS} details Diff-GNSS. Section~\ref{sec:implement} describes datasets, baselines, and training. Section~\ref{sec:experiment} reports comparative and ablation results. Finally, Section~\ref{sec:Conclusion} presents the further discussion, and Section~\ref{sec:Conclusion} concludes.

\section{Related Work}\label{sec:related work}
% \subsection{GNSS Measurement Error Estimation}
% In recent years, numerous studies have sought to enhance GNSS measurement accuracy by detecting anomalous observations or estimating measurement errors. Existing methods fall into two broad categories: traditional approaches and learning-based approaches.

\subsection{Tradition-based Methods for Pseudorange Error Estimation}
% Conventional detection of anomalous GNSS signals primarily relies on measurement-derived cues and environmental priors. Early studies identify anomalies or model errors using raw indicators such as the SNR~\cite{ref06}, ${C/N_{0}}$~\cite{ref07}, and satellite elevation~\cite{ref20}. Considering that multipath or NLOS signals do not always perform consistently with ${C/N_{0}}$ or SNR, leading to high miss and false-alarm rates, particularly in highly reflective settings~\cite{ref21}. Another line of work exploits environmental priors to derive sky masks that identify and exclude multipath/NLOS induced by buildings. A prevalent approach is 3D matching, which assesses satellite signal quality against 3D city models. For example,~\cite{ref22} detects NLOS by leveraging the similarity between building geometry and satellite visibility, whereas~\cite{ref23} employs building boundaries and heights with ray tracing to simulate and correct NLOS effects. Sky masks constructed from LiDAR or fisheye camera can also mitigate many building-induced anomalies~\cite{ref09,ref24}. Building on this,~\cite{ref25} further detects large dynamic objects to suppress their impact. Although these methods improve positioning accuracy~\cite{ref08}, they remain limited by substantial computational cost and the availability of 3D models or auxiliary sensors~\cite{ref26}.
Conventional detection of anomalous GNSS signals uses measurement-derived cues and environmental priors. Early work identifies anomalies with raw indicators such as the SNR~\cite{ref06}, ${C/N_{0}}$~\cite{ref07}, and satellite elevation~\cite{ref20}. Considering that multipath or NLOS signals do not always perform consistently with ${C/N_{0}}$ or SNR, causing high miss and false-alarm rates in highly reflective environments~\cite{ref21}. Another line leverages environmental priors to construct sky masks that identify and remove building-induced multipath/NLOS. A common strategy is 3D matching, evaluating satellite visibility against 3D city models, such as via building-visibility similarity~\cite{ref22} or using building boundaries and heights with ray tracing~\cite{ref23}. Sky masks derived from LiDAR or fisheye camera similarly mitigate many building-caused anomalies~\cite{ref09,ref24}. Extending this idea, large dynamic objects can also be detected and suppressed~\cite{ref25}. Although these methods improve positioning accuracy~\cite{ref08}, they remain limited by substantial computational cost and the availability of 3D models or mauxiliary sensors~\cite{ref26}.

\subsection{Learning-based Methods for Pseudorange Error Estimation}
% Learning-based approaches for detecting anomalous GNSS signals and predicting measurement errors have received substantial attention. Conventional machine-learning techniques show promise for multipath and NLOS detection, common choices include GBDTs and support vector machines (SVMs). For instance,~\cite{ref11} uses GBDT with satellite elevation and pseudorange residuals to predict pseudorange errors, yielding measurable gains in positioning accuracy. Using ${C/N_{0}}$, satellite elevation, and pseudorange residuals as inputs,~\cite{ref27} applies an SVM with a radial-basis-function kernel and achieves high LOS(line-of-sight)/multipath/NLOS classification accuracy.~\cite{ref28} benchmarks GBDT, random forests, decision trees, and k-means, finding high accuracy at sites overlapping the training locations but substantial degradation at unseen sites. Overall, these traditional machine learning-based methods have low computational overhead, but struggle to generalise in complex and changing environments. This is primarily due to their simple model structures being unable to capture the complex, higher-order feature representations in GNSS signal propagation.

Learning-based methods for anomalous GNSS detection and error prediction have received substantial attention. Conventional machine-learning models, such as GBDTs and support vector machines (SVMs), show promise but limited generalization.~\cite{ref11} uses GBDT to predict pseudorange errors from satellite elevation and residuals, improving positioning accuracy. Using ${C/N_{0}}$, elevation, and residuals, an RBF-kernel SVM attains high LOS(line-of-sight)/multipath/NLOS classification accuracy~\cite{ref27}. A broader benchmark of GBDT, random forests, decision trees, and k-means reports strong results at seen sites but marked degradation at unseen ones~\cite{ref28}. Overall, while these methods are computationally efficient, they struggle in complex, changing environments due to their limited capacity to capture higher-order feature representations in GNSS signal propagation.

% The strong ability of deep learning to extract complex representations opens new opportunities for handling anomalous GNSS signals.~\cite{ref29,ref30} use convolutional neural networks for multipath detection and achieve higher accuracy than SVMs.~\cite{ref31} proposes a multipath detection algorithm based on gated recurrent units (GRUs), which improves positioning accuracy by excluding anomalous satellites during localization. Given that satellite signal quality is highly correlated with the environment, the above methods are limited by their lack of explicit environmental representation~\cite{ref26}.~\cite{ref12} utilises FCNNs and LSTMs to learn environment representations from single-epoch GNSS measurements and jointly perform signal classification and pseudorange error estimation. Subsequent work incorporates self-attention to enhance the LSTM's use of context~\cite{ref10}, enabling indirect environment modeling~\cite{ref32}.~\cite{ref26} adopts graph neural networks, allowing the model to flexibly accept inputs with varying numbers and orders of satellites. While these methods achieve impressive accuracy in classifying signals by enriching environmental representations, their prediction of pseudorange errors remains unsatisfactory. This is primarily because pseudorange errors exhibit highly complex distributions~\cite{ref10}.
The capacity of deep learning for complex representation opens new opportunities for handling anomalous GNSS signals. Convolutional neural networks outperform SVMs in multipath detection~\cite{ref29,ref30}.~\cite{ref31} proposes a multipath detector based on gated recurrent units (GRUs), which improves positioning accuracy by excluding anomalous satellites during localization. Yet these methods often lack explicit environmental modeling~\cite{ref26}. To address this,~\cite{ref12} utilises FCNNs and LSTMs to learn environment representations from single-epoch GNSS measurements and jointly perform signal classification and pseudorange error estimation. Subsequent work adds self-attention to better exploit context~\cite{ref10}, enabling indirect environment modeling~\cite{ref32}. Graph neural networks further accommodate varying satellite numbers and orders~\cite{ref26}. Despite improved classification via enriched environment representations, pseudorange error regression remains limited due to its highly complex distributions~\cite{ref10}.
% that current architectures struggle to learn and model.
In this work, we revisit pseudorange error estimation by asking: \textit{How can we model the complex distribution of pseudorange errors within deep learning frameworks to improve estimation accuracy?} To this end, we introduce a strongly conditioned diffusion model that guides denoising to progressively map Gaussian noise to the empirical error distribution.

\subsection{Diffusion Model Application}
Diffusion models have garnered widespread attention owing to their successful applications across diverse domains. They learn complex data distributions by progressively corrupting data with noise and learning the reverse denoising process. Strong performance has been reported in image generation~\cite{ref15,ref33}, video generation~\cite{ref34}, object detection~\cite{ref16}, 3D point-cloud generation~\cite{ref35,ref36} and registration~\cite{ref17,ref37}, scene-flow estimation~\cite{ref18}, and human pose estimation~\cite{ref38}. DifFlow3D~\cite{ref18} proposes a diffusion-based scene-flow estimation network that iteratively refines flow residuals under strong conditions while also modeling uncertainty. PWDLO~\cite{ref37} presents a diffusion-driven hierarchical optimization framework for LiDAR odometry, using a GRU denoiser for fine-grained pose-residual refinement. These advances motivate our work. \textit{However, it remains unclear whether diffusion can effectively model pseudorange errors.}

\section{Diff-GNSS}\label{sec:Diff-GNSS}
\subsection{System Overview}
\begin{figure*}
\centering
\includegraphics[width=0.85\linewidth]{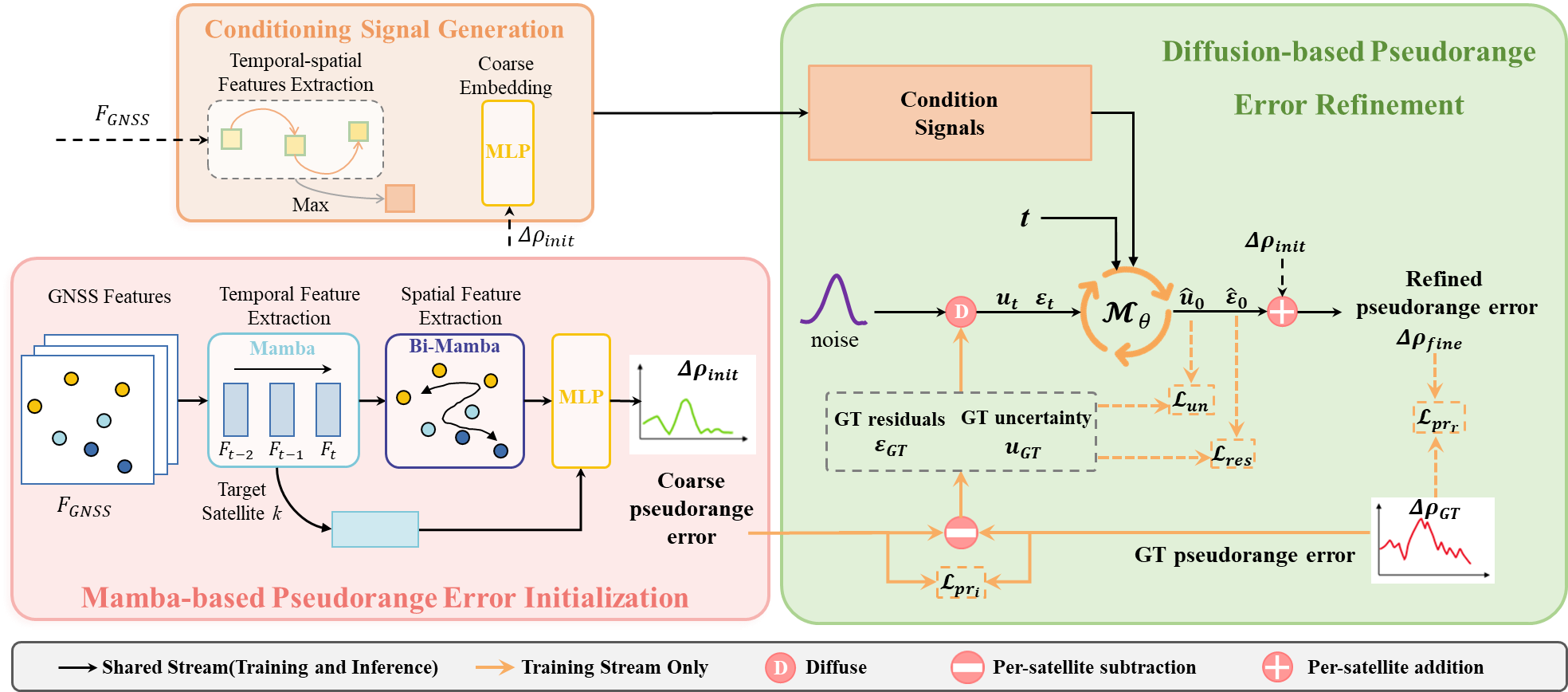}
\vspace{-5pt}
% \caption{Overall architecture of the proposed Diff-GNSS. We first initialize a coarse pseudorange error. Next, temporal-spatial features and a coarse embedding are used as conditioning signals to guide a diffusion-based refinement layer to recover pseudorange residuals. In addition, we jointly predict per-satellite uncertainty to assess the reliability of the pseudorange error estimates.}
\caption{Overall architecture of the proposed Diff-GNSS.}
\label{fig:system}
\end{figure*}

This study proposes Diff-GNSS, a novel diffusion-based coarse-to-fine optimization framework for pseudorange error estimation. The overall architecture is shown in Fig.~\ref{fig:system}. Diff-GNSS takes a five-dimensional per-satellite feature vector as input, comprising the least-squares pseudorange error and its root-sum-of-squares (RSS), ${C/N_{0}}$, satellite elevation, and satellite azimuth. These features are obtained from raw GNSS observations through preprocessing. All visible satellite features within a fixed time window are used as network inputs. The construction of satellite features is described in Section~\ref{sec:GNSS-feature}.

We first perform coarse pseudorange error prediction with Mamba, adopting a two-stage modeling strategy that captures the time-varying characteristics and environmental correlations of pseudorange errors. Specifically, a unidirectional Mamba encodes the GNSS feature sequence along the time dimension to extract per-satellite temporal representations. Subsequently, a bidirectional Mamba (Bi-Mamba) operates along the satellite dimension to mine inter-satellite dependencies from forward and backward context, yielding a global environmental representation. The temporal and spatial features are then concatenated and fed to a multilayer perceptron (MLP) head to produce per-satellite coarse estimates ${\Delta \rho _{init} }$. Details are provided in Section~\ref{sec:Mamba init}.

Next, we refine the pseudorange error using a conditionally guided diffusion model to predict residuals. To evaluate the reliability of per-satellite estimates, we jointly perform uncertainty estimation. In the diffusion-based refinement layer, spatiotemporal features together with a coarse embedding condition a GRU-based denoiser, which iteratively removes noise from both the residual and its uncertainty. The refined pseudorange error ${\Delta \rho _{fine} }$ is obtained by adding the generated pseudorange error residual ${\hat{\varepsilon } _{0} }$ to the coarse prediction ${\Delta \rho _{init} }$. The architecture of this module is detailed in Section~\ref{sec:Diff refine}.

\subsection{GNSS Features Construction}\label{sec:GNSS-feature}
Preprocessing raw GNSS measurements to construct well-founded input features is crucial for filtering out irrelevant information and enhancing the model's sensitivity to key factors. Building on the proven effectiveness of ~\cite{ref10,ref12} for predicting pseudorange error, we construct five GNSS features: satellite elevation, azimuth, ${C/N_{0}}$, least-square pseudorange error, and its RSS, which respectively capture geometric visibility, directional dependence, signal quality, error scale, and environmental complexity. Specifically, least-square pseudorange error and its RSS are constructed as follows:

\subsubsection{Least-Square Pseudorange Error}
The least-square pseudorange error is obtained by first estimating the receiver state via SPP and then, for each satellite, taking the difference between the measured pseudorange and its theoretical value.
% This error reflects the quality of GNSS observations directly and is correlated with the magnitude of the true pseudorange error. This makes it a reliable indicator of the error scale for the network. 
The least-square pseudorange error of the ${k-th}$ satellite ${\Delta \rho _{LS,k}}$ is given by:
\begin{align}
\Delta \rho _{LS,k} =\hat{\rho} _{s_{k},r} -\left \| \mathbf{p} _{s_{k}}-\hat{\mathbf{p} } _{r}   \right \| 
\label{eqn:3}
\end{align}
where, ${\hat{\rho} _{s_{k},r}}$ denotes the pseudorange measurement after correction for atmospheric delays and satellite clock bias. ${\mathbf{p} _{s_{k}}}$ and ${\hat{\mathbf{p} } _{r}}$ denote the satellite position and the receiver position estimated via SPP, respectively.

\subsubsection{Root-Sum-Squares of Pseudorange Error (RSS)}
RSS of pseudorange errors quantifies the overall error level across all visible satellites within the same epoch and provides an effective encoding of environmental context. 
% In general, open-sky scenes produce a low RSS, whereas urban canyons, where multipath and occlusion are more pronounced, produce a higher RSS. Because RSS correlates with external environmental conditions, including it as an input enhances the model's environmental awareness and cross-scene generalization.
The RSS of the ${k-th}$ satellite ${RSS_{k}}$ is computed as:
\begin{align}
    RSS_{k}=\sqrt{\sum_{k=1}^{N}\left ( \Delta \rho _{LS,k} \right )^{T} \cdot \Delta \rho _{LS,k}  } 
    \label{eqn:4}
\end{align}
where, ${N}$ is the total number of satellites received in one epoch.

\subsection{Mamba-based Pseudorange Error Initialization}\label{sec:Mamba init}
Pseudorange errors can span from tens to hundreds of metres, and typically exhibit non-Gaussian, long-tailed distributions under environmental influences. Modeling the full error directly with a diffusion model inflates the generative search space and hampers accuracy. DifFlow3D~\cite{ref18} has demonstrated the effectiveness of a ``coarse estimation - diffusion refinement" strategy for scene-flow estimation. Motivated by this, we insert a coarse prediction stage before diffusion to provide an initial estimate with appropriate scale and trend. This design markedly shrinks the diffusion search space and focuses the denoising on fine-grained residual modeling.

\begin{figure}[!t]
\centering
\vspace{-8pt}
\includegraphics[width=0.6\linewidth]{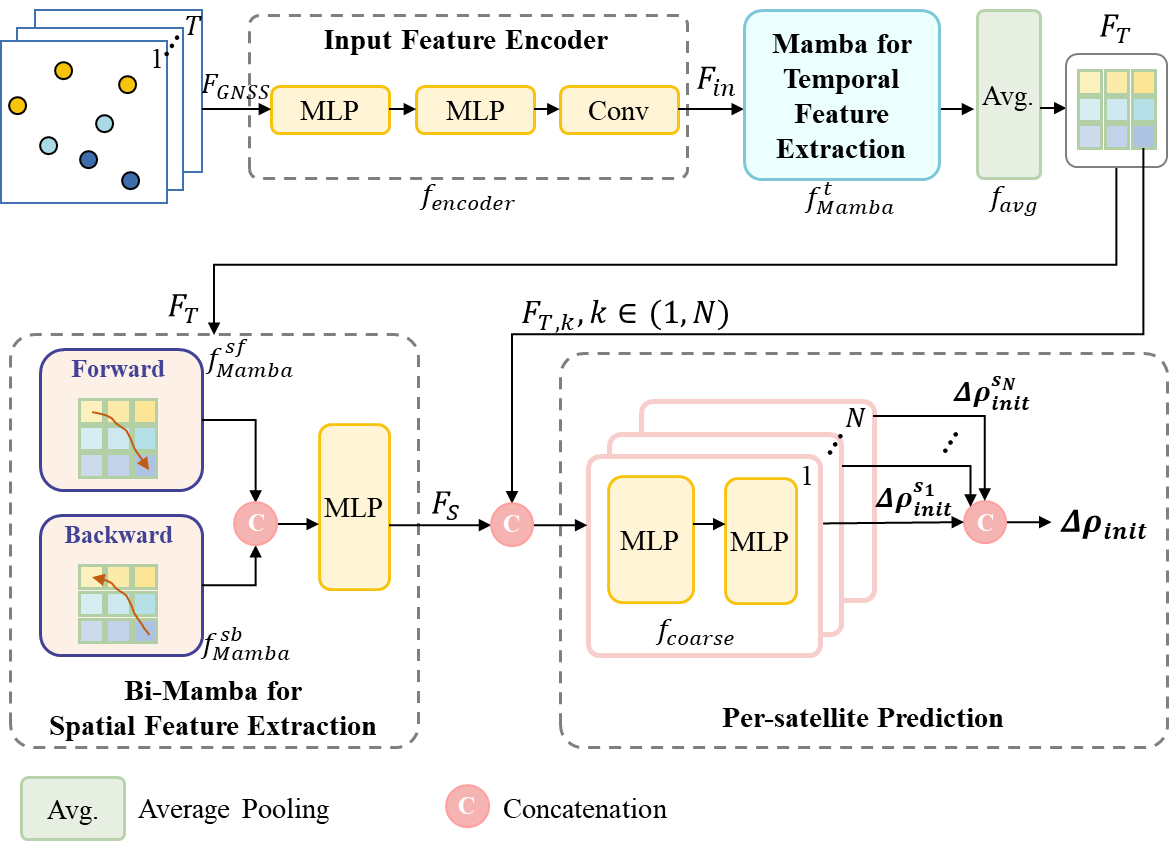}
\vspace{-8pt}
\caption{Architecture of Mamba-based pseudorange error initialization.}
\label{fig:mamba_init}
\vspace{-12pt}
\end{figure}

Mamba~\cite{ref39} fuses the parallel efficiency of state-space models with the the selective representational capacity of attention, showing superior potential to Transformers and RNNs for long-sequence modeling. We therefore design a Mamba-based coarse-prediction module, the full architecture is shown in Fig.~\ref{fig:mamba_init}.

The network takes five-dimensional GNSS features ${F_{GNSS}\in \mathbb{R} ^{N\times T\times D} }$ over ${T}$ time steps and ${N}$ satellites as input. Firstly, two MLPs separately map the feature and temporal channels to higher-dimensional embeddings. A one-dimensional convolution along the temporal dimension. Through this lightweight encoder, we can obtain high-dimensional feature representation ${F_{in}}$:
\begin{align}
    F_{in}=f_{encoder}\left ( F_{GNSS} \right ) 
    \label{eqn:5}
\end{align}

Subsequently, the encoded features are passed to a temporal Mamba block to extract per-satellite pseudorange error dynamics. T To suppress local noise and enhance global perception, we apply average pooling along the temporal dimension, producing the final temporal representation ${F_{T}}$:
\begin{align}
    F_{T}=f_{avg}\left ( f_{Mamba}^{t}\left ( F_{in} \right )   \right )  
    \label{eqn:6}
\end{align}

As pseudorange quality is highly dependent on the environment, we design a Bi-Mamba-based spatial encoder to model the global spatial context. Specifically, the temporal features ${F_{T}}$ are processed by forward ${f_{Mamba}^{sf}}$ and backward ${f_{Mamba}^{sb}}$ Mamba blocks along the satellite dimension to derive bidirectional spatial embeddings. These embeddings are concatenated and passed through an MLP to yield the global spatial representation ${F_{S}}$ for the current epoch. This design captures cross-satellite dependencies arising from observation geometry and directional occlusion, while mitigating sensitivity to input satellite ordering. The construction of the global spatial feature ${F_{S}}$ is formulated as follows:
\begin{align}
    F_{S} = \operatorname{MLP}\bigl( f^{\mathrm{sf}}_{\mathrm{Mamba}}(F_{T}) \;\oplus\; f^{\mathrm{sb}}_{\mathrm{Mamba}}(F_{T}) \bigr)
    \label{eqn:7}
\end{align}

Concatenate the temporal representation ${F_{T}}$ and spatial representation ${F_{S}}$ along the channel dimension to form a spatiotemporal feature for each satellite. Subsequently, a MLP regression head predicts the per-satellite initial pseudorange error ${\Delta \rho _{init}^{s_{k}} }$:
\begin{align}
    \Delta \rho _{init}^{s_{k}} =f_{coarse}\left ( F_{T,k}\oplus F_{S}  \right )  
    \label{eqn:8}
\end{align}

The coarse pseudorange error estimates for ${N}$ satellites can be expressed as ${\Delta \mathbf{\rho}  _{init}  = \left \{ \Delta \rho _{init}^{s_{k}}   \right \} _{k=1}^{N}  }$.

\subsection{Diffusion-based Pseudorange Error Refinement}\label{sec:Diff refine}
To obtain more accurate pseudorange error estimates, we introduce a diffusion-based pseudorange error refinement model that iteratively generates pseudorange error residuals to refine the coarse prediction. The overall pipeline is shown in Fig.~\ref{fig:system}. During training, the forward diffusion process progressively adds Gaussian noise to the ground-truth (GT) residuals and their uncertainties to construct noisy samples. Subsequently, the condition guides the denoising network to learn the inverse denoising process on the noisy samples. At inference, under the same conditioning, the model starts from Gaussian noise and iteratively denoises to obtain accurate residual estimates, enabling fine-grained correction of pseudorange errors. The involved modules are detailed below.

\subsubsection{Forward Diffusion Process}
A diffusion model is a generative framework comprising a forward diffusion and a reverse denoising process: Data are progressively converted into Gaussian noise, and the reverse process is learnt to approximate the data distribution. Specifically, the forward process injects Gaussian noise into GT samples at time steps ${(1,\dots ,T)}$ to produce noisy samples. In our pseudorange error refinement task, the GT samples are the true pseudorange error residuals ${\varepsilon _{GT}^{s_{i}} }$:
\begin{align}
    \varepsilon _{GT}^{s_{i}} = \varepsilon _{0}^{s_{i}}=\Delta \rho _{GT}^{s_{i}}- \Delta \rho _{init}^{s_{i}}  
    \label{eqn:9}
\end{align}
where, ${ \varepsilon _{0}^{s_{i}}}$ denotes the ground-truth pseudorange error residual of the ${i-th}$ satellite, ${\Delta \rho _{GT}^{s_{i}}}$ represents the GT pseudorange error of the ${i-th}$ satellite, and ${\Delta \rho _{init}^{s_{i}}}$ denotes the coarse pseudorange error estimate. The progressive injection of Gaussian noise is modeled as a Markov process:
\begin{align}
    q\left ( \varepsilon _{1:T}^{s_{i}}\mid \varepsilon _{0}^{s_{i}}   \right ) =\prod_{t=1}^{T} q\left ( \varepsilon _{t}^{s_{i}}\mid \varepsilon _{t-1}^{s_{i}}   \right )  
    \label{eqn:10}
\end{align}
where, ${t\in \left ( 1,\dots ,T \right ) }$ is the sampling step, ${\varepsilon _{t}^{s_{i}}}$ is the noisy pseudorange error residual of the ${i-th}$ satellite at sampling step ${t}$. ${q\left ( \varepsilon _{t}^{s_{i}}\mid \varepsilon _{t-1}^{s_{i}}   \right )}$ denotes the Gaussian transition kernel, which can be written as:
\begin{align}
    q\left ( \varepsilon _{t}^{s_{i}}\mid \varepsilon _{t-1}^{s_{i}}   \right )=\mathcal{N} \left (  \varepsilon _{t}^{s_{i}};\sqrt{1-\beta _{t} } \cdot \varepsilon _{t-1}^{s_{i}},\beta _{t}\cdot \mathbf{I}   \right )    
    \label{eqn:11}
\end{align}
where, ${\beta _{t}\in [0,1)}$ is hyperparameter, ${\mathbf{I}}$ is the identity matrix. Based on the incremental noising mechanism in (\ref{eqn:10}), the process of generating the pseudorange error residual at an arbitrary sampling step ${\varepsilon _{t}^{s_{i}}}$ from ${\varepsilon _{0}^{s_{i}}}$ can be expressed as:
\begin{align}
    \varepsilon _{t}^{s_{i}}=\sqrt{\alpha _{t} } \varepsilon _{0}^{s_{i}}+\sqrt{1-\alpha _{t}}Z,Z\sim \mathcal{N}(0,\mathbf{I} )      
    \label{eqn:12}
\end{align}
where, ${\alpha _{t}= {\textstyle \prod_{k=1}^{t}}\left ( 1-\beta _{k}  \right ) }$ controls the intensity of the injected Gaussian noise. When ${T}$ is large enough, the GT samples degenerate into pure Gaussian noise.

\subsubsection{Reverse Denoising Process}
The reverse denoising process trains a neural network ${\mathcal{M}_{\theta } \left ( \varepsilon _{t}^{s_{i}},t;C  \right ) }$ to iteratively denoise the noisy input ${\varepsilon _{t}^{s_{i}}}$, yielding the denoised sample ${\hat{\varepsilon}  _{0}^{s_{i}}}$. Within a Markov-chain formulation, this process can be expressed as:
\begin{align}
    \varepsilon _{t-1}^{s_{i}}=&\sqrt{\alpha_{t-1} }\mathcal{M}_{\theta } \left ( \varepsilon _{t}^{s_{i}},t;C  \right )+\frac{\sqrt{1-\alpha _{t-1}-\sigma _{t}^{2} } }{\sqrt{1-\alpha _{t}} } \notag \\
&\left ( \varepsilon _{t}^{s_{i}}-\sqrt{\alpha _{t}}\mathcal{M}_{\theta }\left ( \varepsilon _{t}^{s_{i}},t;C  \right )   \right )   +\sigma _{t}Z       
    \label{eqn:13}
\end{align}
where, ${C}$ represents the conditioning signal. ${\sigma _{t} }$ denotes the covariance of the Gaussian distribution at time step ${t}$.

\subsubsection{Uncertainty Estimation}
% The reliability of perception network outputs is critical, especially under varying input conditions, where the confidence of the estimated pseudorange errors may differ. Accordingly, we explicitly incorporate uncertainty modeling into the diffusion refinement head.
In order to model the uncertainty associated with each satellite during the diffusion-based optimization process, thereby achieving a more accurate estimation, we have designed an uncertainty module.
As illustrated in Fig.~\ref{fig:system}, at each diffusion step the network not only estimates the pseudorange error residual for each satellite but also predicts a corresponding uncertainty to quantify confidence in the estimate. This uncertainty is supervised using a carefully defined GT uncertainty ${u_{GT}}$, which is calculated as follows:
\begin{align}
    e_{ab}^{s_{i}} &=\left | \Delta \rho _{init}^{s_{i}} -\Delta \rho _{GT}^{s_{i}} \right | , \notag \\
e_{re}^{s_{i}} &=\left | \frac{e_{ab}^{s_{i}}}{\left | \Delta \rho _{GT}^{s_{i}}  \right | }  \right |, \notag \\
u_{GT}^{s_{i}} &=  u_{0}^{s_{i}}=\left\{\begin{matrix}0,
  &e_{ab}^{s_{i}}< E_{1},e_{re}^{s_{i}}< E_{2} \\1,
  &otherwise
\end{matrix}\right.
\label{eqn:14}
\end{align}
where, ${e_{ab}^{s_{i}}}$ denotes the absolute deviation between the coarse pseudorange error estimate and the GT, ${e_{re}^{s_{i}}}$ quantifies the relative proportion of this absolute deviation with respect to the GT pseudorange error. The absolute and relative deviation thresholds are specified by ${E_{1}}$ and ${E_{2}}$, respectively. Additionally, we adopt the same noise strategy and noise addition method (\ref{eqn:12}) as for the pseudorange error when adding noise to the uncertainty.

\subsubsection{Conditioning Signal Generation}
\begin{figure}[!t]
\centering
\includegraphics[width=0.75\linewidth]{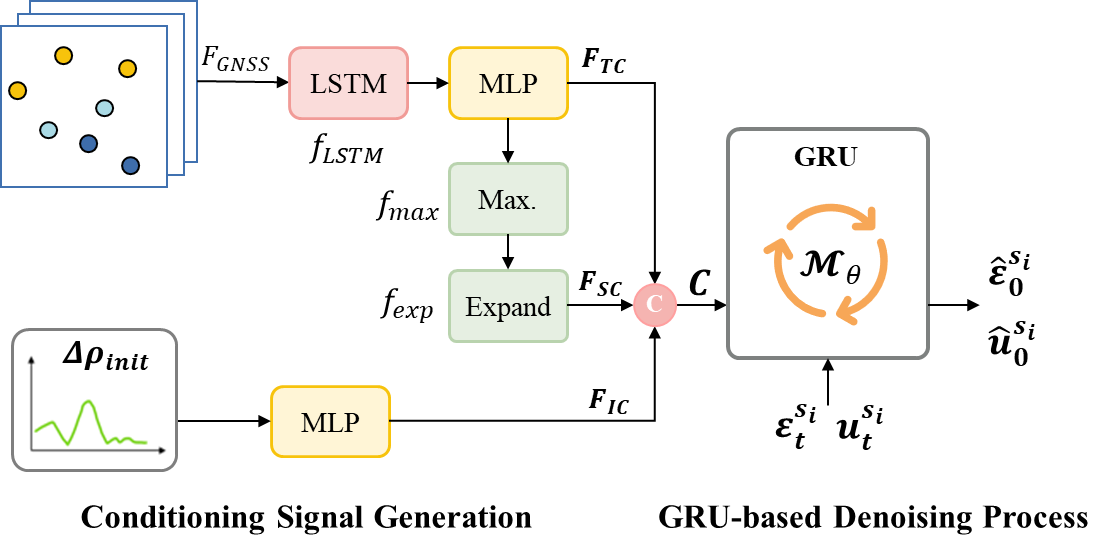}
\vspace{-8pt}
\caption{Architecture of conditioning signal generation and denoising network.}
\vspace{-12pt}
\label{fig:gru}
\end{figure}

Due to their generative diversity, diffusion models are not easily applied directly to deterministic per-satellite pseudorang error regression, as this task requires precise, single-valued estimates. We therefore enforce controllable generation via strong conditioning. Specifically, we construct a conditioning signal that fuses temporal features, spatial context, and the coarse estimate to guide the reverse denoising process; the detailed architecture is shown in Fig.~\ref{fig:gru}. Firstly, we encode GNSS features over time and use an LSTM followed by an MLP to obtain temporal embeddings ${F_{TC}}$:
\begin{align}
F_{TC}=MLP\left ( f_{LSTM}(F_{GNSS})   \right ) 
\label{eqn:15}
\end{align}

We next perform max pooling ${f_{max}}$ across the satellite dimension to aggregate features into a scene-level context vector, which is then replicate along the satellite dimension to align with the number of satellites to furnish a scene-aware representation for each satellite. The complete spatial context is defined as follows:
\begin{align}
F_{SC}=f_{exp}\left ( f_{max}(F_{TC})   \right ) 
\label{eqn:16}
\end{align}

Meanwhile, the coarse prediction ${\Delta \rho _{init}}$ is embedded via an MLP to obtain ${F_{IC}}$, which is then concatenated with ${F_{TC}}$ and ${F_{SC}}$ to form the conditioning signal ${C}$:
\begin{align}
C=F_{IC}\oplus F_{TC}\oplus F_{SC} 
\label{eqn:17}
\end{align}

\subsubsection{Denoisng Network}
After constructing the conditioning signal, we adopt a GRU-based denoiser, following PWDLO~\cite{ref37}, to jointly recover the pseudorange error residual and its uncertainty. The architecture is shown in Fig.~\ref{fig:gru}. Specifically, the noisy samples ${\varepsilon _{t}^{s_{i}} }$ and ${u_{t}^{s_{i}} }$ are concatenated and encoded to initialize the hidden state of the GRU module. The conditioning signal ${C}$ serves as the input sequence that guides iterative updates of the hidden state. At sampling step ${t}$, the network outputs the denoised residual ${\hat{\varepsilon}  _{0}^{s_{i}} }$ together with its uncertainty ${\hat{u}  _{0}^{s_{i}} }$. The overall procedure is formulated as:
\begin{align}
    \hat{\varepsilon}  _{0}^{s_{i}} ,\hat{u}  _{0}^{s_{i}} =\mathcal{M}  _{\theta } \left ( \varepsilon _{t}^{s_{i}},u _{t}^{s_{i}},t,C \right ) 
    \label{eqn:18}
\end{align}

By adding the residual estimate ${\hat{\varepsilon}  _{0}^{s_{i}}}$ from the reverse denoising to the coarse estimate ${\Delta \rho _{init}^{s_{i}}}$, we obtain the refined pseudorange error of the ${i-th}$ satellite ${\Delta \rho _{fine}^{s_{i}}}$:
\begin{align}
    \Delta \rho _{fine}^{s_{i}} =\Delta \rho _{init}^{s_{i}}+\hat{\varepsilon}  _{0}^{s_{i}}
    \label{eqn:19}
\end{align}

\subsection{Training Objective}\label{sec:train object}
To enable end-to-end joint optimization of the Mamba-based coarse estimator and the diffusion-based refiner, we impose hierarchical supervision at three levels, includeing pseudorange error, pseudorange residual, and uncertainty. This yields a multi-scale training objective. As illustrated in Fig.~\ref{fig:system}, the total loss ${\mathcal{L}}$ comprises four terms: the coarse prediction loss ${\mathcal{L}_{pr_{i}}}$, the residual loss ${\mathcal{L}_{res}}$, the uncertainty loss ${\mathcal{L}_{un}}$, and the full pseudorange error loss ${\mathcal{L}_{pr_{r}}}$. The final objective is given by:
\begin{align}
    \mathcal{L} =\lambda_{pr_{i}} \mathcal{L}_{pr_{i}} + \lambda_{res} \mathcal{L}_{res}
+ \lambda_{un} \mathcal{L}_{un} +\lambda_{pr_{r}} \mathcal{L}_{pr_{r}}
\label{eqn:20}
\end{align}
where, ${\lambda_{pr_{i}}}$, ${\lambda_{res}}$, ${\lambda_{un}}$, and ${\lambda_{pr_{r}}}$ represent the weighting coefficient for each loss item. The definitions of each loss item are as follows:
\begin{align}
    \mathcal{L}_{pr_{i}} &=\frac{1}{N} \sum_{i=1}^{N} \left ( \Delta \rho _{init}^{i}- \Delta \rho _{GT}^{i} \right )^{2}, \notag \\
\mathcal{L}_{res} &=\frac{1}{N} \sum_{i=1}^{N} \left ( \varepsilon  _{GT}^{i}- \hat{\varepsilon} _{0}^{i} \right )^{2}, \notag \\
\mathcal{L}_{un} &=\frac{1}{N} \sum_{i=1}^{N} \left ( u  _{GT}^{i}- \hat{u} _{0}^{i} \right )^{2}, \notag \\
\mathcal{L}_{pr_{r}} &=\frac{1}{N} \sum_{i=1}^{N} \left ( \Delta \rho _{fine}^{i}- \Delta \rho _{GT}^{i} \right )^{2}.
\label{eqn:21}
\end{align}

\section{IMPLEMENTATION}\label{sec:implement}
\subsection{Dataset}\label{sec:dataset}
\begin{figure}[!b]
\centering
\vspace{-14pt}
\subfloat[Trajectory of the HK dataset.\label{fig:hkdata}]{
  \includegraphics[width=0.46\linewidth]{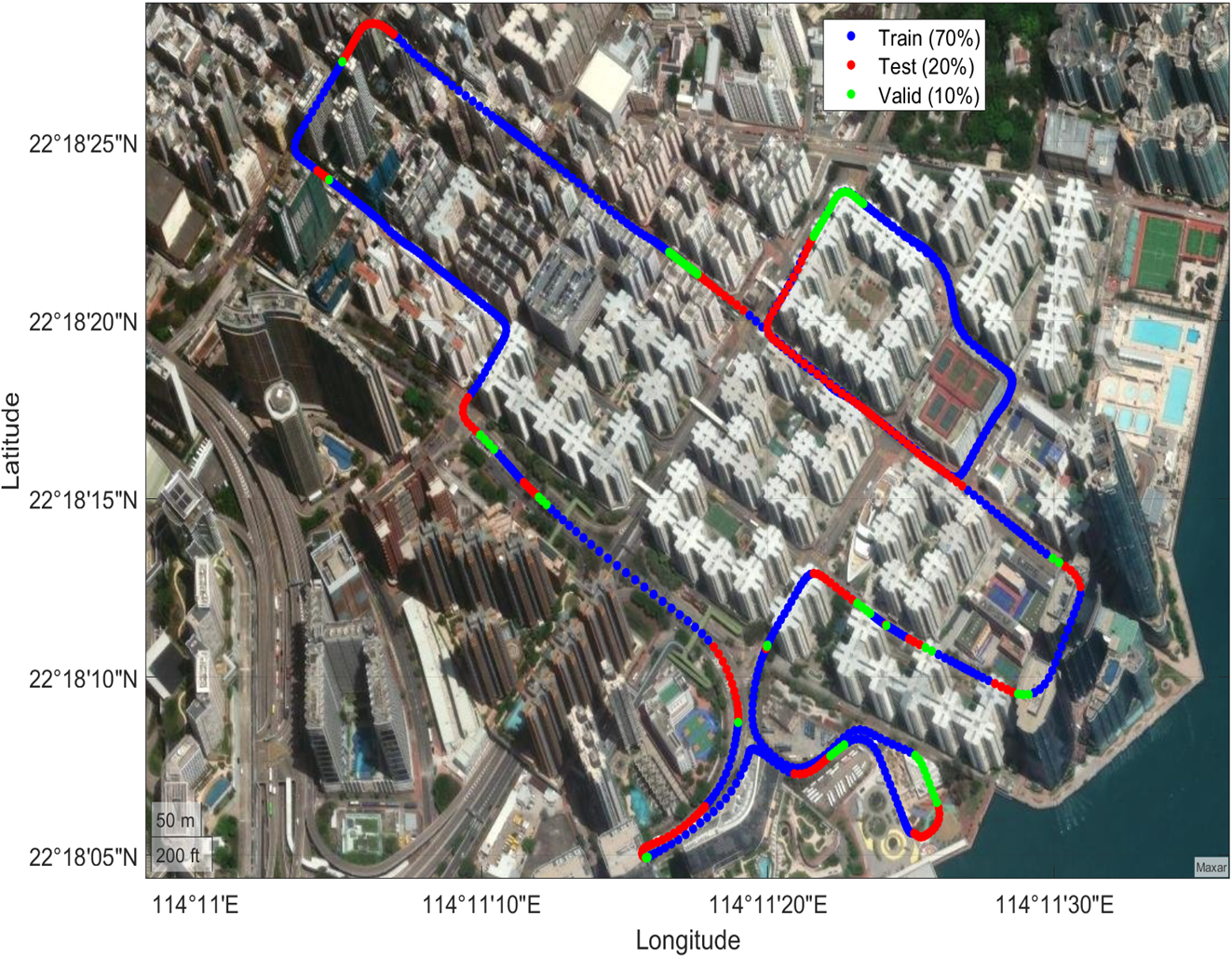}
}\hfil
\subfloat[Trajectory of the in-house dataset.\label{fig:ourdata}]{
  \includegraphics[width=0.46\linewidth]{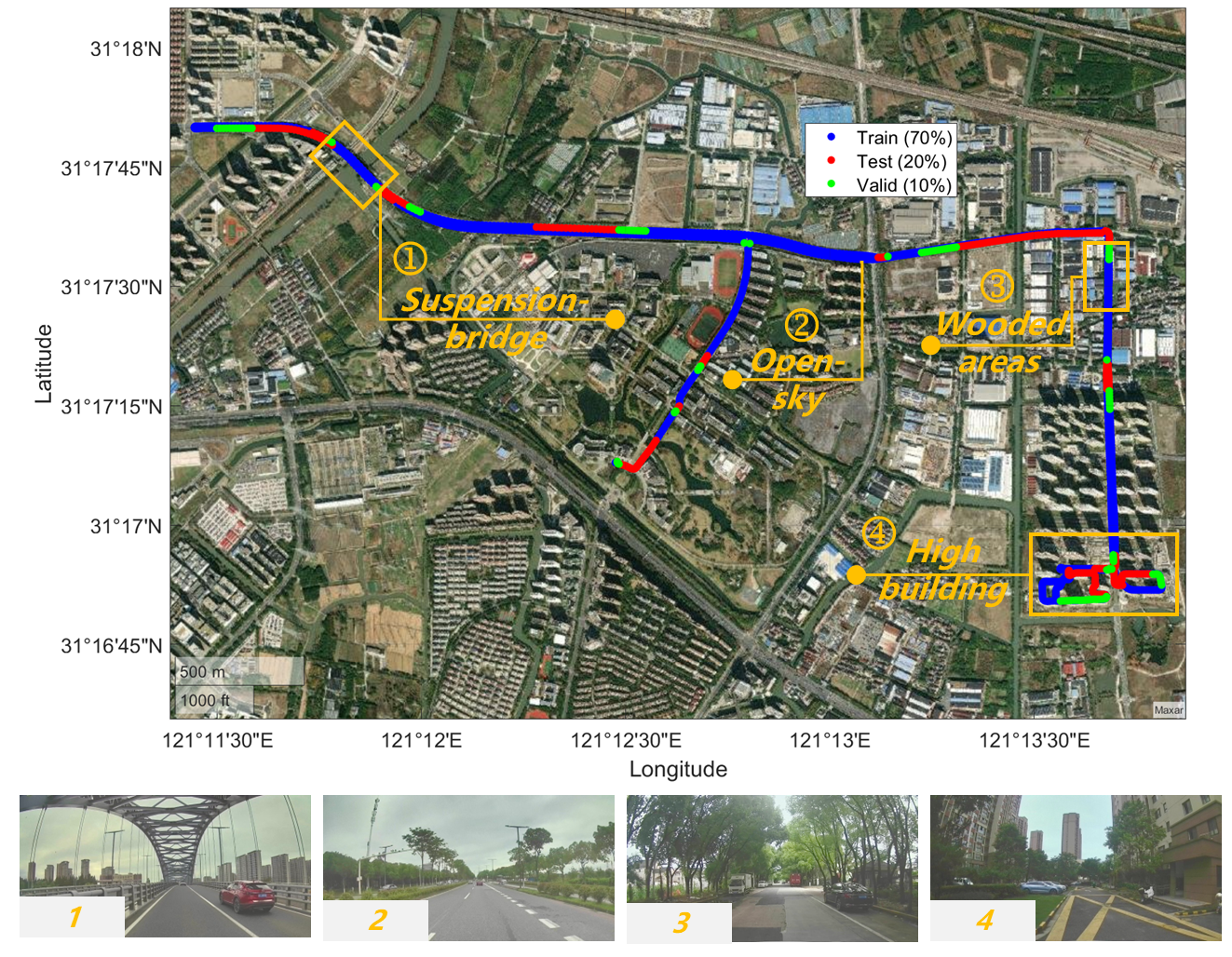}
}
\caption{Trajectories of the two datasets used in the experiments.}
\label{fig:dataset}
\end{figure}

To evaluate the performance of the proposed Diff-GNSS, we conduct experiments using both the public Hong Kong (HK) dataset~\cite{ref12} and a real-world dataset that we have collected ourselves in Shanghai. The trajectories for both datasets are shown in Fig.~\ref{fig:dataset}. The HK dataset is recorded in highly urbanized areas dominated by high-rise buildings, where GNSS measurements are frequently corrupted, yielding large errors. To increase scene diversity and assess generalization, we collect our dataset using a Jiangling Yi platform across regions encompassing open-sky, wooded areas, high buildings, and suspension-bridge scenarios. The test vehicle is equipped with a u-blox F9P receiver, which outputs raw GNSS measurements at a frequency of 1 Hz. GT trajectories are produced by post-processing with IE software, utilising a high-precision GNSS/INS integrated positioning device GJ-VSMS2000.

%要想实现并排放，得调整图片宽度到能并排放下2张图片即可，否则自动将第2张图片放下一排了
% \begin{figure}[t]
% \centering
% \captionsetup[subfigure]{font=scriptsize}
% \subfloat[Trajectory of the HK dataset.\label{fig:hkdata}]{
%   \includegraphics[width=0.9\linewidth]{Figure/HKdataset.png}
% }\hfill
% \subfloat[Trajectory of our dataset.\label{fig:ourdata}]{
%   \includegraphics[width=0.9\linewidth]{Figure/Ourdataset.png}
% }
% \caption{Trajectories of the two datasets used in our experiments.}
% \label{fig:dataset}
% \end{figure}

To account for the strong scene dependence of GNSS measurement quality, we stratify both datasets by scenario. Within each partitioned segment, the data are split into training, validation, and test sets at a 7:1:2 ratio. The partitioning results are shown in Fig.~\ref{fig:dataset}. We use a temporal sliding window of length 3 on the segmented sequences as the network input. Detailed statistics for both datasets are provided in Table~\ref{tab:1}.
\subsection{Data Augmentation}
Considering that this limited sample size is insufficient for effective feature learning, we apply data augmentation to both the training and validation sets. To maintain temporal consistency, we first extract 5${s}$ temporal clips from the raw data. Within each clip, we enumerate all possible sub-sequences of length 3 in chronological order. After removing duplicates, we obtain the augmented dataset. Through data augmentation, we have increased the training sets of the HK and our datasets by 2679 and 7090 frames, respectively. This provided the network with more diverse input data.

\begin{table}[!t]
    \centering
    % \vspace{-8pt}
    \caption{Number of data samples in each dataset.}
    \renewcommand{\arraystretch}{1.2} % 调整行高为原来的1.5倍
    \setlength{\tabcolsep}{12pt}
    \begin{tabular}{c|c|c|c|c}
    \Xhline{1.1pt} % 顶部加粗
    Dataset & Duration & Train & Valid & Test \\ \hline
    HK      & 1534 ${s}$   & 555  & 60   & 141  \\ \hline
    Our     & 2127 ${s}$   & 1450  & 173   & 385  \\ 
    \Xhline{1.1pt} % 底部加粗
    \end{tabular}
    \vspace{-8pt}
    \label{tab:1}
\end{table}

\subsection{Evaluation Details}
To comprehensively evaluate the performance of the proposed Diff-GNSS in pseudorange error estimation, we compare it with SOTA deep-learning baselines: the LSTM-based method, PolyU~\cite{ref12} and the Transformer-enhanced LSTM model, RWTH~\cite{ref10}. The evaluation metrics comprise mean absolute error (MAE), and root mean square error (RMSE) of the pseudorange error estimates.

Both Diff-GNSS and RWTH are trained and evaluated on the datasets constructed in Section~\ref{sec:dataset} with a temporal window of 3. Because PolyU accepts only single-epoch inputs, we construct a dataset with a temporal window of 1 for that method.

\subsection{Training Details}
All training and evaluation are conducted on an NVIDIA GeForce RTX 4090 GPU using PyTorch 1.8.1. We adopt the Adam optimizer with an initial learning rate of 0.00002. Training runs for 200 epochs, and the learning rate is multiplied by 0.9 every 5 epochs. The batch size is 8. We adopt DDIM~\cite{ref40} with a sampling step size of 2 and a total of T=1000 diffusion steps. For (\ref{eqn:20}), ${\lambda _{pr_{i}}=0.5, \lambda _{res}=0.5, \lambda _{un}=0.3,\lambda _{pr_{r}}=1.0}$.

\section{EXPERIMENTAL RESULTS}\label{sec:experiment}
This section reports comparisons between the proposed diffusion-based pseudorange error predictor and SOTA baselines. We then present extensive ablations to quantify the contribution of each component. Finally, we investigate how Diff-GNSS benefits downstream GNSS positioning.

\subsection{Comparison Experiment}
Table~\ref{tab:2} shows the performance comparison with the PolyU and RWTH. Our method achieves the best results across all metrics for both the HK and Our datasets, thus demonstrating its effectiveness. Compared to PolyU, which relies only on single-epoch features, RWTH improves accuracy by introducing self-attention and explicitly modeling temporal dependencies. However, as a deterministic regression approach, RWTH struggles to capture the complex error distributions induced by multipath and NLOS effects, limiting its improvements. Relative to RWTH, our method reduces MAE by 1.24 ${m}$ and 1.04 ${m}$ on HK and our dataset, respectively, and lowers RMSE by 36.3\% and 59.1\%. These improvements demonstrate that, through being trained with progressive noise addition, conditional-diffusion based residual refinement can effectively learn complex distributions and substantially enhance pseudorange error estimation.

\begin{table}[!t]
    \centering
    \caption{The comparison results of the pseudorange error estimates on the HK dataset and our dataset. The best result is highlighted in bold.}
    \renewcommand{\arraystretch}{1.2} % 调整行高为原来的1.5倍
    \setlength{\tabcolsep}{8pt}
    \begin{tabular}{c|c|c|c|c}
    \Xhline{1.1pt} % 顶部加粗
    \multirow{2}{*}{Methods} & \multicolumn{2}{c|}{HK dataset}                    & \multicolumn{2}{c}{Our dataset}                    \\ \cline{2-5} 
                         & \multicolumn{1}{c|}{MAE (${m}$)}       & RMSE (${m}$)      & \multicolumn{1}{c|}{MAE (${m}$)}       & RMSE (${m}$)      \\ \hline
PolyU                    & \multicolumn{1}{c|}{4.03}          & 6.96          & \multicolumn{1}{c|}{2.14}          & 4.71          \\ \hline
RWTH                     & \multicolumn{1}{c|}{3.45}          & 6.33          & \multicolumn{1}{c|}{1.73}          & 4.62          \\ \hline
Ours                     & \multicolumn{1}{c|}{\textbf{2.21}} & \textbf{4.03} & \multicolumn{1}{c|}{\textbf{0.69}} & \textbf{1.89} \\
    \Xhline{1.1pt} % 底部加粗
    \end{tabular}
    \label{tab:2}
\end{table}

\begin{figure}[!t]
\centering
\vspace{-12pt}
\subfloat[Pseudorange error estimation.\label{fig:prerror_HK:a}]{
  \includegraphics[width=0.46\linewidth]{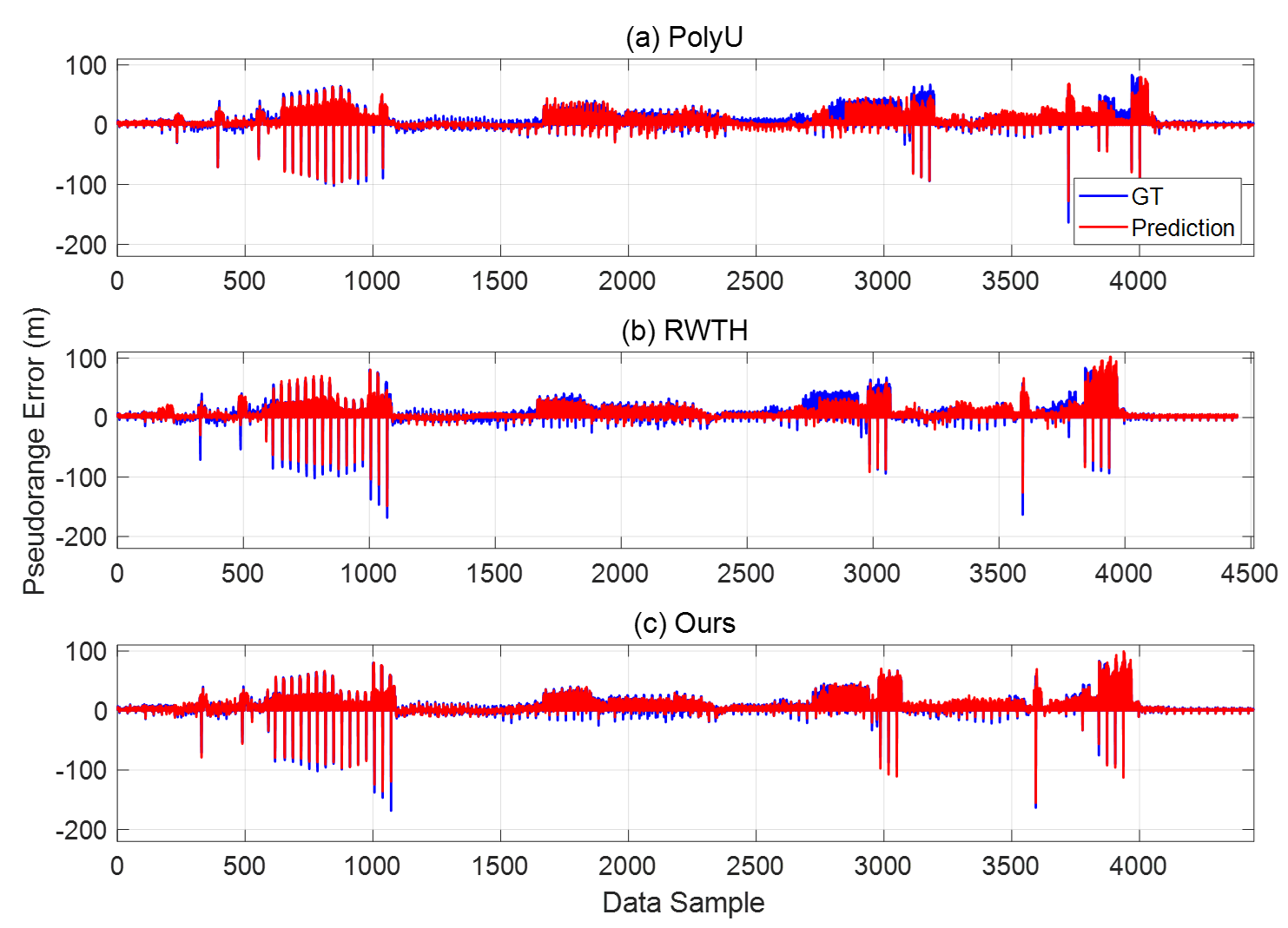}
}\hfil
\subfloat[Prediction errors.\label{fig:prederror_HK:b}]{
  \includegraphics[width=0.46\linewidth]{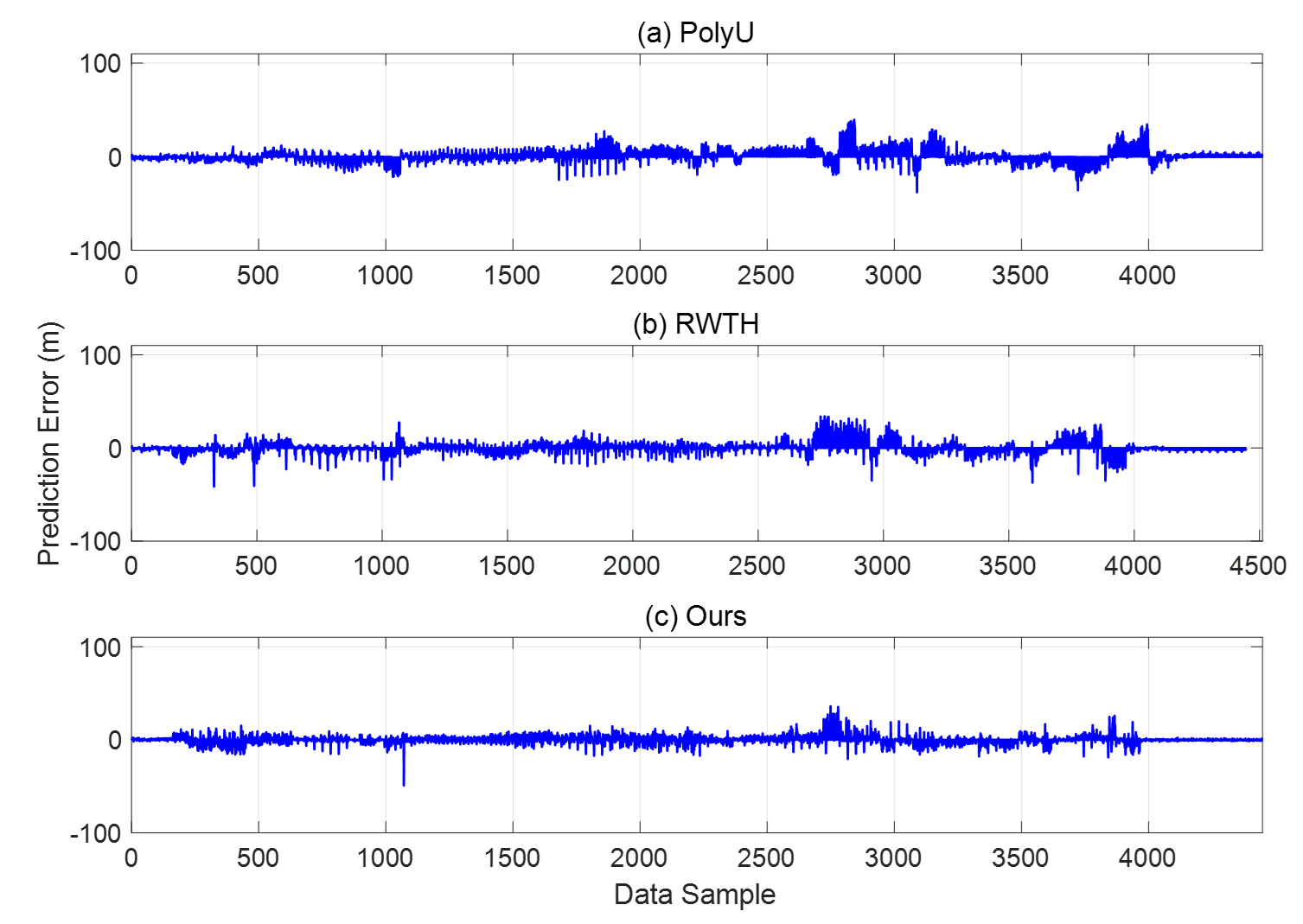}
}
\caption{Pseudorange error estimation and prediction errors on the HK dataset for the proposed method, PolyU, and RWTH.}
\vspace{-10pt}
\label{fig:result_HK}
\end{figure}

\begin{figure}[!t]
\centering
\vspace{-12pt}
\subfloat[Pseudorange error estimation.\label{fig:prerr_our:a}]{
  \includegraphics[width=0.46\linewidth]{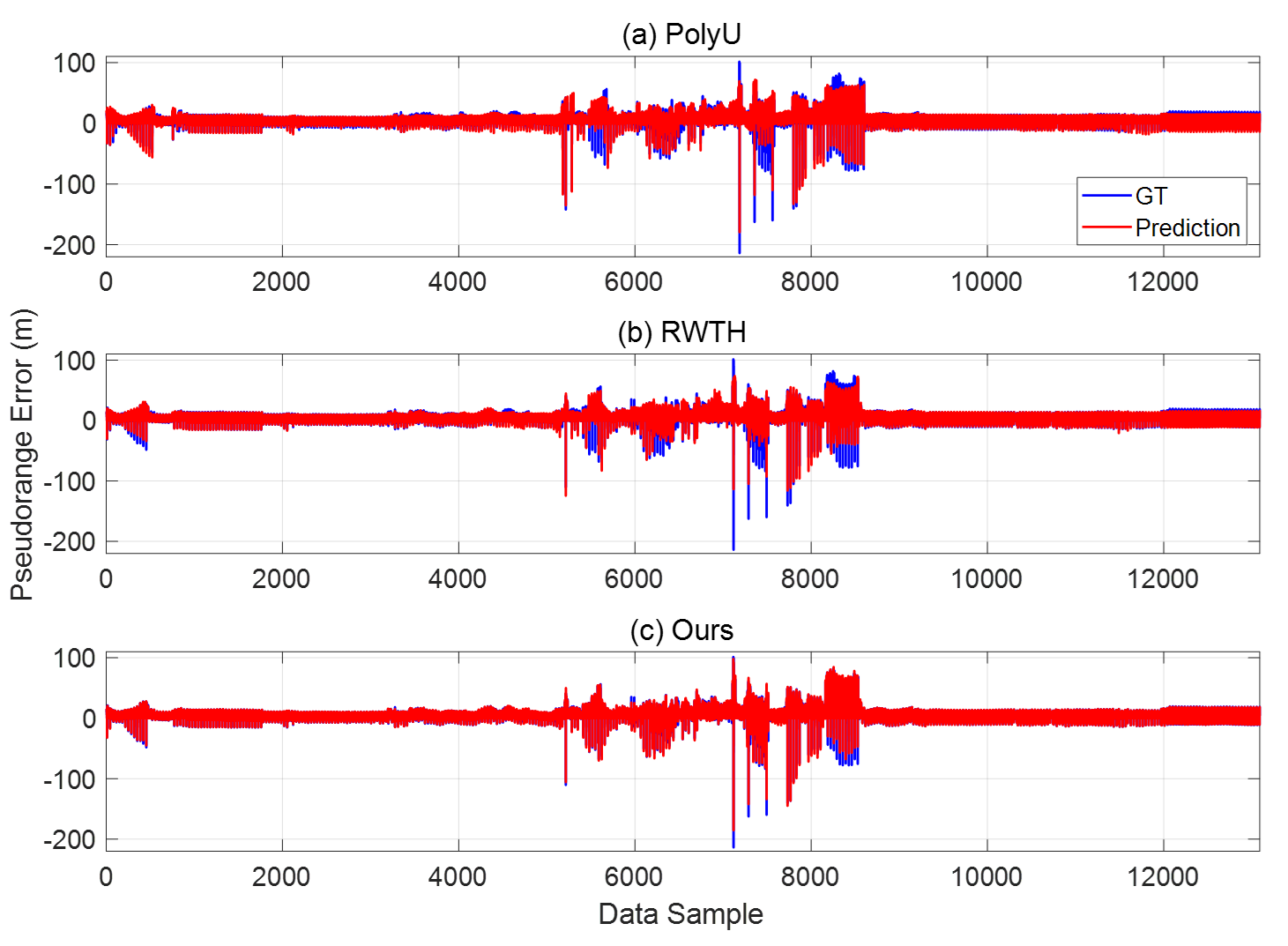}
}\hfil
\subfloat[Prediction errors.\label{fig:prederror_our:b}]{
  \includegraphics[width=0.46\linewidth]{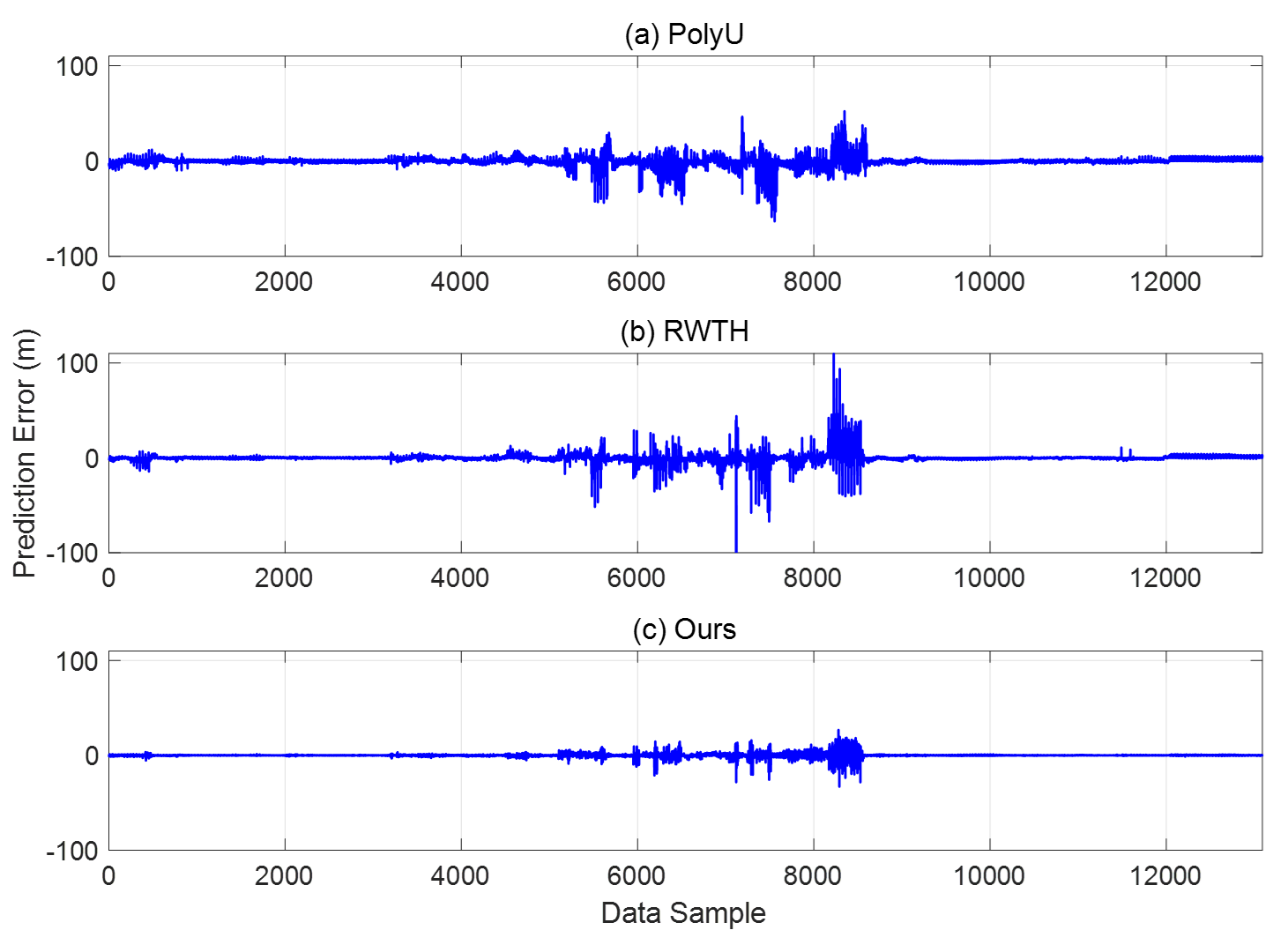}
}
\caption{Pseudorange error estimation and prediction errors on our dataset for the proposed method, PolyU, and RWTH.}
\vspace{-8pt}
\label{fig:result_our}
\end{figure}

Fig.~\ref{fig:result_HK} and~\ref{fig:result_our} provide qualitative evaluations of pseudorange error estimation. Specifically, Fig.~\ref{fig:result_HK} presents the estimated pseudorange errors and the corresponding prediction errors on HK dataset. Compared with the baselines, the proposed method achieves the best estimation accuracy owing to its fine-grained modeling of pseudorange errors. Our dataset covers complex and diverse environments, including urban buildings, wooded areas and suspension bridges, which place greater demands on robustness and generalization. As shown in Fig.~\ref{fig:result_our}, thanks to the superior error-modeling capability of the diffusion-based refiner, our method still attains the best performance.

% \begin{figure}[!t]
% \centering
% \includegraphics[width=1.0\linewidth]{Figure/Prerror_our.png}
% \caption{Results of pseudorange error estimation for the proposed method, PolyU, and RWTH on our dataset.}
% \label{fig:prerror_our}
% \end{figure}

% \begin{figure}[!t]
% \centering
% \includegraphics[width=1.0\linewidth]{Figure/Prederror_our.png}
% \caption{Prediction errors of the proposed method, PolyU, and RWTH on our dataset.}
% \label{fig:prederror_our}
% \end{figure}

% \begin{figure}[!t]
% \centering
% \includegraphics[width=1.0\linewidth]{Figure/Prerror_HK.png}
% \caption{Results of pseudorange error estimation for the proposed method, PolyU, and RWTH on HK dataset.}
% \label{fig:prerror_HK}
% \end{figure}

% \begin{figure}[!t]
% \centering
% \includegraphics[width=1.0\linewidth]{Figure/Prederror_HK.png}
% \caption{Prediction errors of the proposed method, PolyU, and RWTH on HK dataset.}
% \label{fig:prederror_HK}
% \end{figure}

\begin{figure}[!t]
\centering
% \vspace{-8pt}
\includegraphics[width=0.6\linewidth]{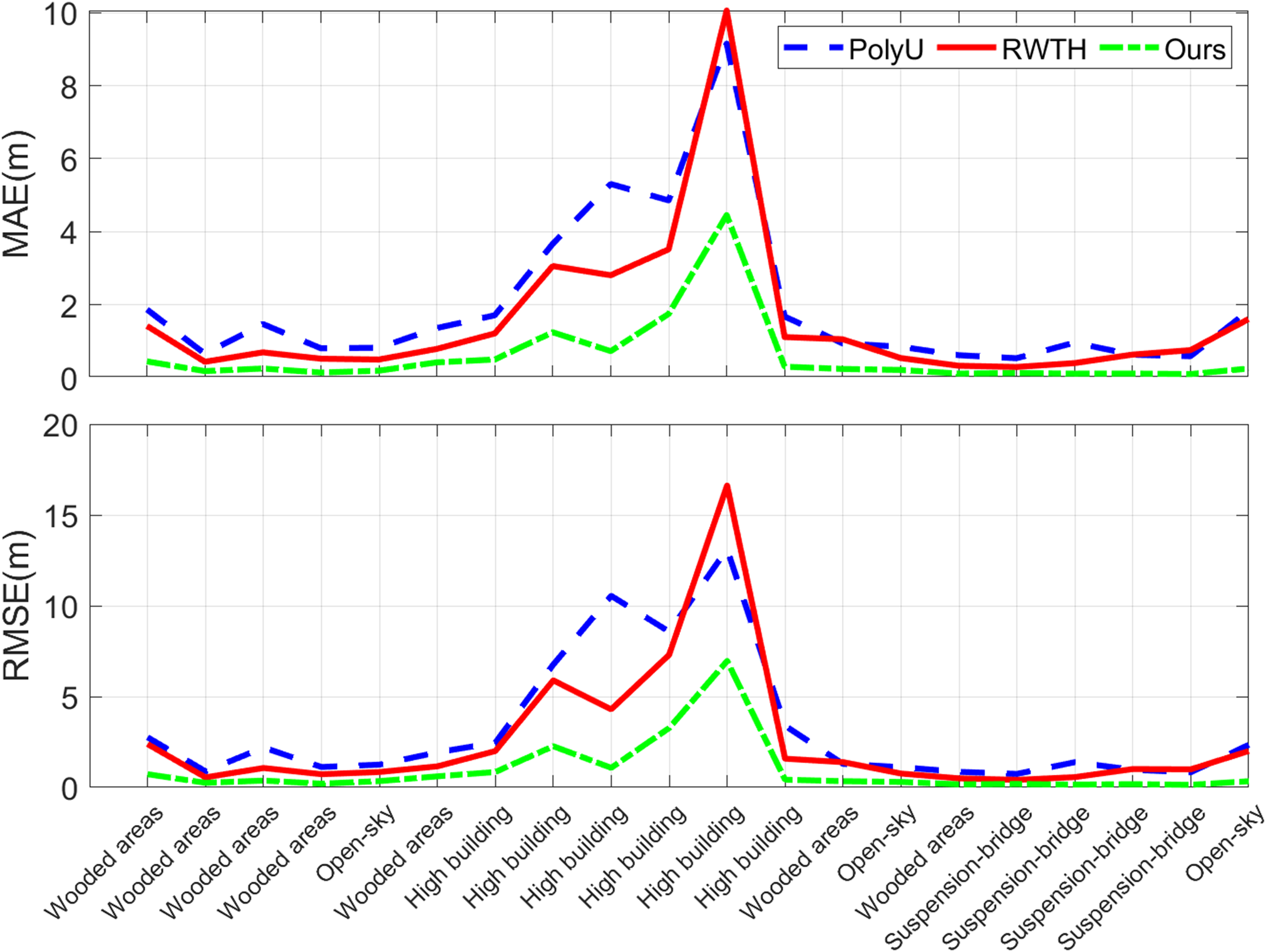}
\vspace{-8pt}
\caption{Performance of the proposed method, PolyU, and RWTH across different scenarios.}
\label{fig:performance_scene}
\end{figure}

To evaluate performance across different scenes, we conduct a scene-wise comparison on our dataset, which includes multiple scenario types. The results are shown in Fig.~\ref{fig:performance_scene}. Overall, our method achieves the best performance in all scenes. In open-sky, wooded areas, and suspension-bridge scenarios, all three methods perform comparably with high accuracy. However, in high-rise scenarios, strong multipath and occlusion cause intermittent satellite outages and disrupt temporal continuity, resulting in significantly larger errors. Despite this challenge, our approach still achieves substantially lower MAE and RMSE than the baselines.

\begin{figure}[!t]
\centering
\vspace{-12pt}
\subfloat[GPS satellites.\label{fig:prerror_gps:a}]{
  \includegraphics[width=0.46\linewidth]{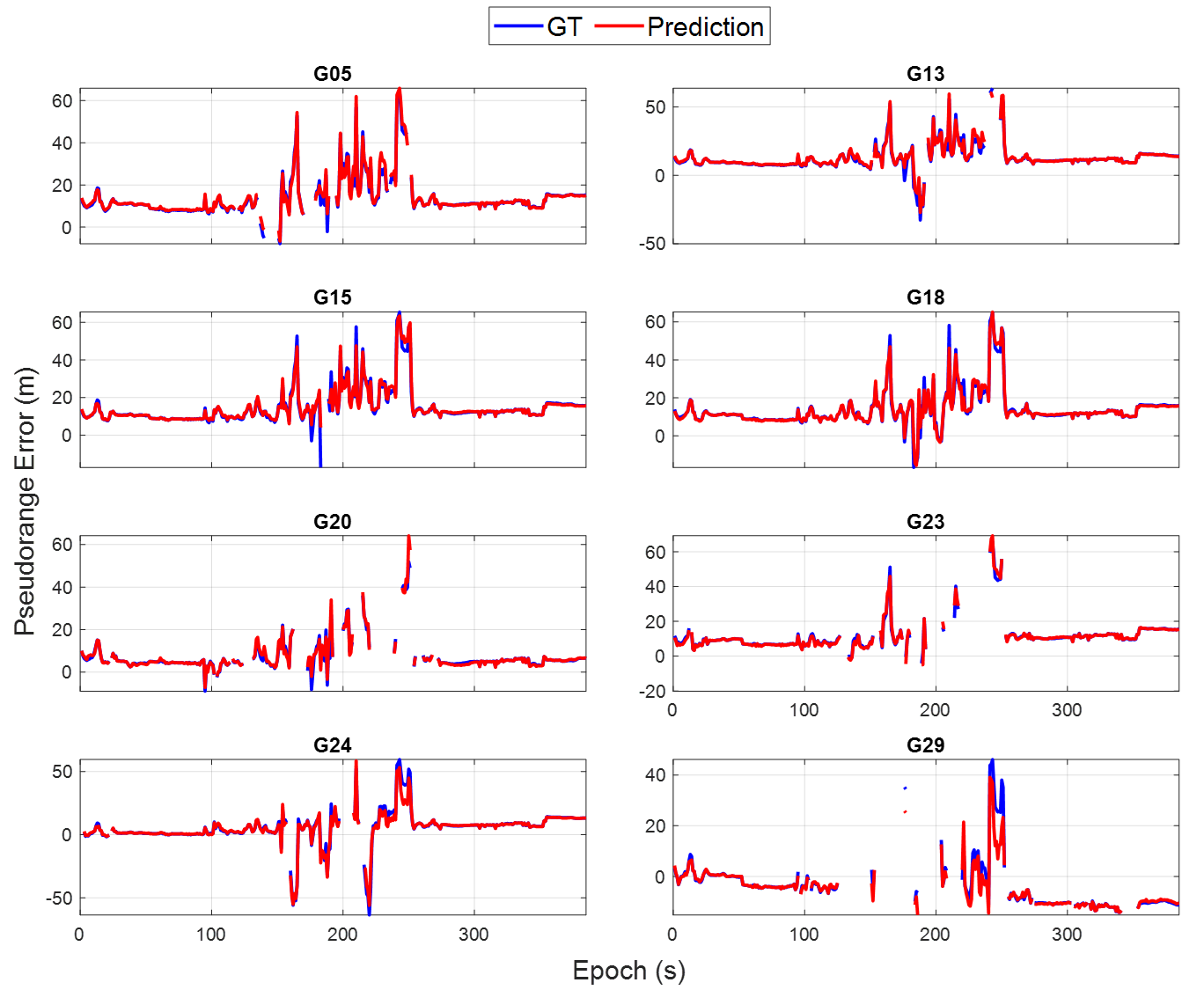}
}\hfil
\subfloat[BDS satellites.\label{fig:prerror_bds:b}]{
  \includegraphics[width=0.46\linewidth]{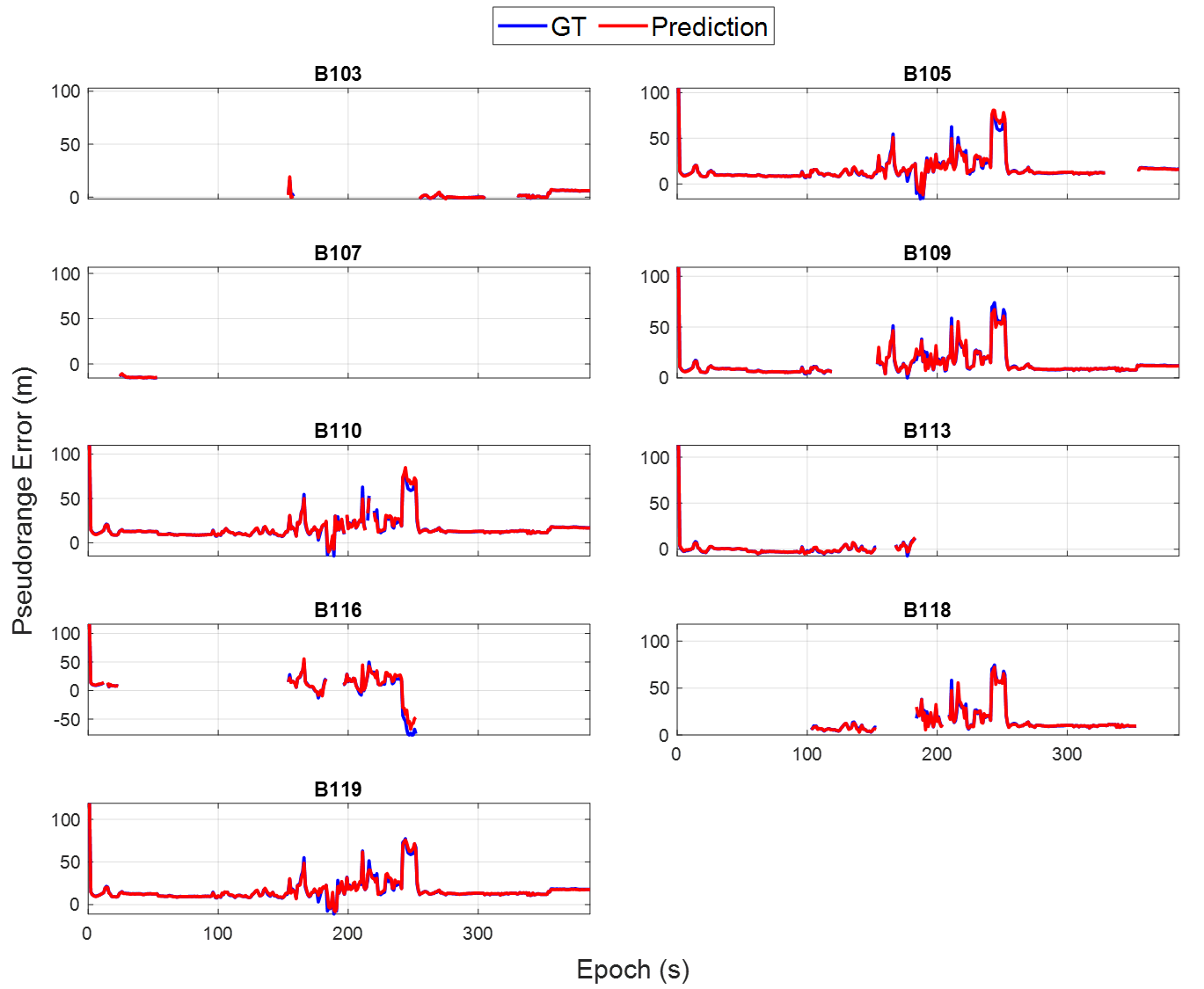}
}
\caption{Pseudorange error estimation for GPS and BDS satellites on the in-house dataset using the proposed method.}
\vspace{-12pt}
\label{fig:prerr_gps_bds}
\end{figure}

Fig.~\ref{fig:prerr_gps_bds} depicts per-satellite pseudorange error estimates for the GPS and BDS constellations. As our data is collected through dynamic testing in a variety of typical scenes, including open-sky, wooded areas and high-rise areas, the pseudorange error of the same satellite can vary substantially across scenes. Nevertheless, our method produces estimates whose magnitudes closely match the GT. Not only does the model accurately track abrupt surges of more than 40 m within seconds, it also detects rapid recoveries to nominal levels. This demonstrates strong modeling of intra-satellite temporal structure. In addition, performance is also consistent across satellites, indicating effective adaptation to inter-satellite distributional shifts.

% \begin{figure}[!t]
% \centering
% \includegraphics[width=1.0\linewidth]{Figure/GPS.png}
% \caption{Results of pseudorange error estimation for GPS satellites using the proposed method on our dataset.}
% \label{fig:prerror_gps}
% \end{figure}

% \begin{figure}[!t]
% \centering
% \includegraphics[width=1.0\linewidth]{Figure/BDS.png}
% \caption{Results of pseudorange error estimation for BDS satellites using the proposed method on our dataset.}
% \label{fig:prerror_bds}
% \end{figure}

\subsection{Ablation Study}\label{sec:Ablation Study}
We have conducted extensive ablation studies to validate the effectiveness of every proposed module. The results are presented in Table~\ref{tab:3}.

\begin{table}[!t]
    \centering
    % \vspace{-8pt}
    \caption{The ablation study results of pseudorange error estimation on our dataset. The best result is highlighted in bold.}
    \renewcommand{\arraystretch}{1.0}
    \setlength{\tabcolsep}{8pt}
    \begin{tabular}{l|cc} % 仅在第一列右侧保留竖线
    \Xhline{1.1pt}
Methods                                                                                                          & MAE (m)       & RMSE (m)      \\ \hline
\multicolumn{1}{l|}{\cellcolor[HTML]{EFEFEF}(a) Diffusion}                                                                                  & \multicolumn{2}{c}{}          \\
Ours (w/o diffusion)                                                                                             & 1.18          & 3.21          \\
Ours (full)                                                                                                      & \textbf{0.69} & \textbf{1.89} \\ \hline
\multicolumn{1}{l|}{\cellcolor[HTML]{EFEFEF}(b) Condition}                                                                                & \multicolumn{2}{c}{}          \\
Ours (w/o temporal feature)                                                                                      & 0.71          & 2.18          \\
Ours (w/o spatial feature)                                                                                       & 0.70          & 2.14          \\
Ours (w/o coarse embedding)                                                                                      & 0.73          & 2.30          \\
Ours (full)                                                                                                      & \textbf{0.69} & \textbf{1.89} \\ \hline
\multicolumn{1}{l|}{\cellcolor[HTML]{EFEFEF}(c) Uncertainty}                                                                             & \multicolumn{2}{c}{}          \\
Ours (w/o uncertainty)                                                                                           & 0.73          & 2.24          \\
Ours (full)                                                                                                      & \textbf{0.69} & \textbf{1.89} \\ \hline
\multicolumn{1}{l|}{\cellcolor[HTML]{EFEFEF}\begin{tabular}[c]{@{}l@{}}(d) Mamba-based \\ Pseudorange Error Initialization\end{tabular}} & \multicolumn{2}{c}{}          \\
Ours (with LSTM)                                                                                                 & 1.22          & 2.36          \\
Ours (with Transformer)                                                                                          & 1.18          & 2.86          \\
Ours (with unidirectional Mamba)                                                                                 & 0.70          & 2.13          \\
Ours (full)                                                                                                      & \textbf{0.69} & \textbf{1.89} \\
    \Xhline{1.1pt}
    \end{tabular}
    \vspace{-8pt}
    \label{tab:3}
\end{table}

\subsubsection{Effectiveness of diffusion model}
We first remove the diffusion-based refinement module, leaving only the coarse pseudorange error predictor. As shown in Table~\ref{tab:3}(a), the full model significantly outperforms the coarse predictions, underscoring the effectiveness of diffusion-based refinement for pseudorange error estimation.

\subsubsection{Impact of Condition Signals}
Conditioning signals play a crucial guiding role in the diffusion-based pseudorange error refinement module. We evaluate their roles through leave-one-out ablations, and the results are shown in Table~\ref{tab:3}(b). When the temporal or spatial feature conditions are removed, performance of our method degrades: relative to the full model, RMSE increases by 15.34\% and 13.23\%, respectively. This is because spatiotemporal features provide effective cross-time dependencies and environmental context. In the absence of the coarse embedding, performance deteriorates markedly: MAE and RMSE rise by 5.8\% and 21.69\%, respectively. These findings indicate that the coarse pseudorange error estimate provides a strong guidance signal for residual generation.

\subsubsection{Impact of Uncertainty}
\begin{figure}[!t]
\centering
% \vspace{-8pt}
\includegraphics[width=0.6\linewidth]{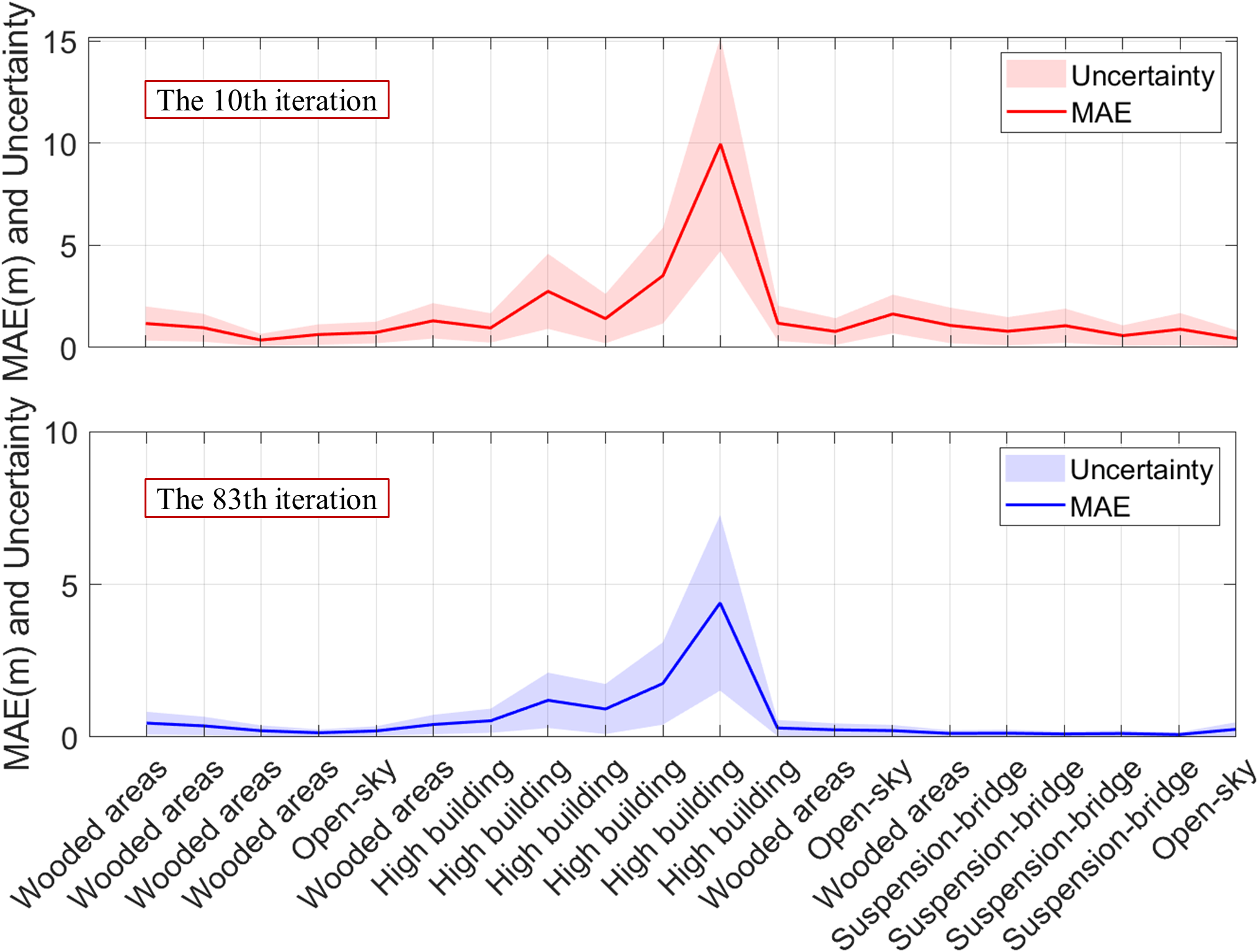}
\vspace{-8pt}
\caption{Results of MAE and prediction uncertainty at different iteration counts.}
\vspace{-14pt}
\label{fig:uncertainty}
\end{figure}

Table~\ref{tab:3} shows that removing the proposed per-satellite uncertainty estimator increases MAE and RMSE to 0.73${m}$ and 2.24${m}$, respectively. This indicates that being aware of uncertainty benefits pseudorange error estimation. To further illustrate its role, Fig.~\ref{fig:uncertainty} plots MAE versus the predicted uncertainty across different iteration counts. As the number of iterations grows, the uncertainty intervals contract markedly while MAE decreases accordingly. A larger MAE indicates greater uncertainty. This indicates the model's ability to capture uncertainty. These observations suggest that uncertainty modeling can effectively guide the model toward more reliable predictions.

\subsubsection{Impact of Mamba-based Pseudorange Error Initialization}
To quantify the contribution of the proposed Mamba-based pseudorange error initialization module, we perform ablations under identical training and evaluation settings by replacing all Mamba blocks with LSTM or Transformer counterparts. The results are shown in Table~\ref{tab:3}(d). The Mamba variant achieves the best results across all metrics. This indicates that its selective state space modelling can reliably capture long-range dependencies, outperforming RNN-based and attention-based models when processing long sequences with heteroscedastic noise.

Furthermore, to verify the effectiveness of the proposed Bi-Mamba for spatial feature extraction, we replace it with a unidirectional variant. The bidirectional design consistently outperforms the unidirectional one across metrics, likely because it aggregates information from both preceding and succeeding satellites, thereby reducing sensitivity to the input satellite ordering.

\subsection{Runtime Analysis}
% Table~\ref{tab:4} summarizes the runtimes for the proposed Diff-GNSS and baseline methods, which were all evaluated under identical configurations.
We have summarizes the runtime of the proposed Diff-GNSS and baseline methods, with all methods evaluated under identical configurations. The results are as follows: PolyU: 5.5 $ms$, RWTH: 17.5 $ms$, Diff-GNSS: 7.3 $ms$. 
Compared to the baselines, Diff-GNSS adds only a modest amount of computational overhead, yet it achieves larger accuracy gains, providing a favourable accuracy-efficiency trade-off. Notably, RWTH incurs higher computational cost owing to its Transformer component. Moreover, because the GNSS data are sampled at 1 Hz, the 7.3 ${ms}$ inference time is well below the 1000 ${ms}$ budget and thus satisfies real-time processing.

% \begin{table}[!t]
%     \centering
%     \caption{Runtime comparison.}
%     \renewcommand{\arraystretch}{1.2} % 调整行高为原来的1.5倍
%     \setlength{\tabcolsep}{14pt}
%     \begin{tabular}{c|c|c|c}
%     \Xhline{1.1pt} % 顶部加粗
%     Methods      & PolyU & RWTH & Ours \\ \hline
% Runtime ($ms$) & 5.5   & 17.5 & 7.3  \\ 
%     \Xhline{1.1pt} % 底部加粗
%     \end{tabular}
%     \label{tab:4}
% \end{table}

\section{Further Discussion}

\subsection{Plug-and-play on Existing Works}
Notably, the proposed diffusion-based refinement module is plug-and-play: integrating it into several baselines yields significant accuracy gains. To verify this property, we adopt PolyU and RWTH as coarse pseudorange error estimation branches and append the diffusion refinement module afterward, with no architectural changes and no additional hyperparameter tuning apart from minimal interface adaptation. Then, we jointly train the complete model. As shown in Table~\ref{tab:5}, adding diffusion refinement reduces MAE by 36.45\% and 26.01\% for the two methods, respectively, with corresponding improvements in RMSE.

\begin{table}[!t]
    \centering
    % \vspace{-12pt}
    \caption{The plug-and-play capability of our methods. Our Diffusion-based Pseudorange Error Refinement (DPER) can effectively improve the accuracy introduced into recent methods on our datasets.}
    \renewcommand{\arraystretch}{1.2} % 调整行高为原来的1.5倍
    \setlength{\tabcolsep}{12pt}
    \begin{tabular}{c|c|c}
    \Xhline{1.1pt} % 顶部加粗
    Methods    & MAE (${m}$)                 & RMSE (${m}$)                \\ \hline
PolyU      & 2.14                    & 4.71                    \\ \hline
PolyU+DPER & \textbf{1.36 {\color[HTML]{32CB00} ($\downarrow$ 36.45\%)}} & \textbf{3.62 {\color[HTML]{32CB00} ($\downarrow$ 23.14\%)}} \\ \hline
RWTH       & 1.73                    & 4.62                    \\ \hline
RWTH+DPER  & \textbf{1.28 {\color[HTML]{32CB00} ($\downarrow$ 26.01\%)}} & \textbf{3.77 {\color[HTML]{32CB00} ($\downarrow$ 18.40\%)}} \\
    \Xhline{1.1pt} % 底部加粗
    \end{tabular}
    \label{tab:5}
\end{table}

\subsection{Impact of Positioning}
Pseudorange measurements are the main observable for GNSS single-point positioning (SPP). Accurately estimating and compensating their errors can substantially improve positioning accuracy. To quantify the benefit, we evaluate two methods on our dataset within a factor-graph framework: SPP with raw pseudoranges (``SPP-raw"), and SPP with pseudoranges corrected by the proposed method (``SPP-correction"), while keeping all other settings fixed. Both GPS and BDS constellations are considered.

\begin{figure}[!t]
\centering
\vspace{-8pt}
\includegraphics[width=0.6\linewidth]{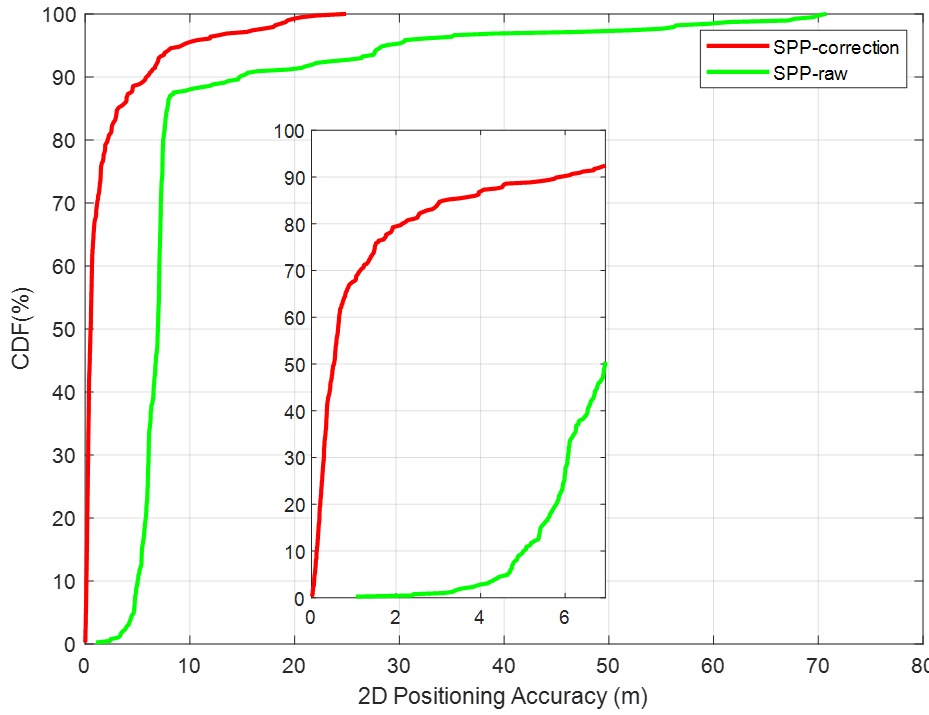}
\vspace{-8pt}
\caption{Results of the availability of 2D positioning.}
\vspace{-8pt}
\label{fig:CDF}
\end{figure}

As shown in Table~\ref{tab:6}, correcting the pseudoranges yields significant reductions in the mean horizontal positioning error and in the RMSE along the east, north, and up axes. This indicates that the proposed estimator provides higher-quality observations to the back end and improves positioning accuracy. We also provide a comparison of the availability of 2D positioning using cumulative distribution functions (CDFs) in Fig.~\ref{fig:CDF}, which further demonstrates the significant benefits of accurate pseudorange error correction.

\begin{table}[!t]
    \centering
    % \vspace{-8pt}
    \caption{Results of the positioning.}
    \renewcommand{\arraystretch}{1.2} % 调整行高为原来的1.5倍
    \setlength{\tabcolsep}{8pt}
    \begin{tabular}{c|c|c}
    \Xhline{1.1pt} % 顶部加粗
    Methods        & \begin{tabular}[c]{@{}c@{}}Mean horizontal \\ positioning error (m)\end{tabular}          & RMSE of E/N/U (m)                                     \\ \hline
SPP-raw        & 9.62                                           & 5.64/ 13.28/ 26.67                                      \\ \hline
SPP-correction & {\textbf{1.96 {\color[HTML]{32CB00}($\downarrow$ 79.60\%)}}} & \textbf{\begin{tabular}[c]{@{}c@{}}2.48 {\color[HTML]{32CB00}($\downarrow$ 56.03\%)}/\\ 3.51 {\color[HTML]{32CB00}($\downarrow$ 73.57\%)}/\\ 4.27 {\color[HTML]{32CB00}($\downarrow$ 83.37\%)}\end{tabular}} \\
    \Xhline{1.1pt} % 底部加粗
    \end{tabular}
    \label{tab:6}
\end{table}

\section{Conclusion}\label{sec:Conclusion}
This paper proposes Diff-GNSS, a novel diffusion-based framework for estimating GNSS pseudorange errors, which adopts a coarse-to-fine strategy for fine-grained modelling. The coarse module employs Mamba to jointly model temporal dynamics and spatial correlations, yielding an efficient initialization. The diffusion-based refinement module uses spatiotemporal features and a coarse embedding as strong conditioning signals to guide a GRU-based denoiser to generate pseudorange error residuals. In addition, we jointly predict per-satellite uncertainty, endowing the model with uncertainty awareness. Extensive experiments demonstrate the effectiveness of Diff-GNSS and significantly improve positioning accuracy. To the best of our knowledge, this is the first application of diffusion models to pseudorange error estimation. Moreover, the diffusion-based refinement module is plug-and-play, providing a new paradigm for future research. In the future, we will explore tight integration with factor-graph frameworks to develop an end-to-end pseudorange-based positioning system.

% References

\bibliographystyle{IEEEtranTIE}
\bibliography{diff_gnss_newRefs}
% \bibliography{Bibliography/IEEEabrv,Bibliography/BIB_xx-TIE-xxxx}\ 
%IEEEabrv instead of IEEEfull

%\vspace{-1cm}
% \begin{IEEEbiography}
% [{\includegraphics[width=1in,height=1.25in,clip,keepaspectratio]{Figure/证件照zhu.png}}]{Jiaqi Zhu} received the B.E. degree in vehicle engineering from the School of Automotive Studies, Jilin University, Jilin, China, in 2019. She is currently pursuing the Ph.D. degree in vehicle engineering with the School of Automotive Studies, Tongji University, Shanghai, China. 

% Her research interests include state estimation, multi-sensor fusion, monocular visual inertial odometry, and GNSS precise positioning.
% \end{IEEEbiography}

\vspace{-30pt}
\begin{IEEEbiography}[{\includegraphics[width=1in,height=1.25in,clip,keepaspectratio]{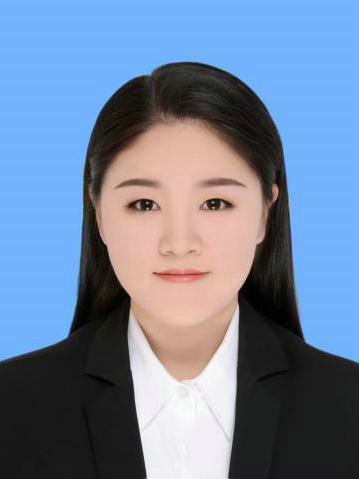}}]{Jiaqi Zhu}
received the B.E. degree in vehicle engineering from the School of Automotive Studies, Jilin University, Jilin, China, in 2019. She is currently pursuing the Ph.D. degree in vehicle engineering with the School of Automotive Studies, Tongji University, Shanghai, China. 

Her research interests include state estimation, multi-sensor fusion, monocular visual-inertial odometry, and GNSS precise positioning.
\end{IEEEbiography}

\vspace{-20pt}
\begin{IEEEbiography}[{\includegraphics[width=1in,height=1.25in,clip,keepaspectratio]{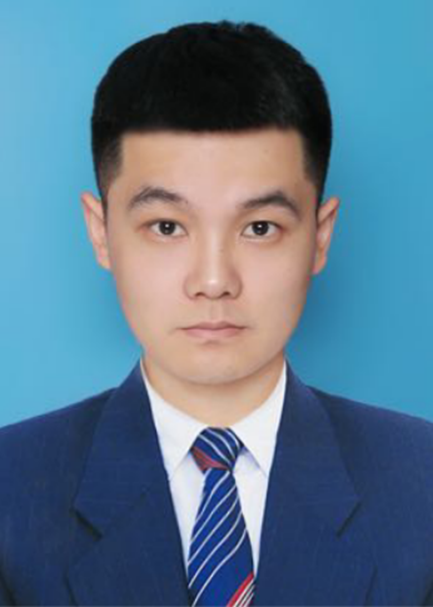}}]{Shouyi Lu}
received the B.S. degree from Shandong University of Technology in 2019 and the M.S. degree from Jilin University in 2022, all in vehicle engineering. He is currently pursuing the Ph.D. degree in vehicle engineering at the School of Automotive Studies, Tongji University, Shanghai, China.

His research interests include SLAM and computer vision, multi-sensor fusion, radar signal processing, and deep learing.
\end{IEEEbiography}

\vspace{-20pt}
\begin{IEEEbiography}[{\includegraphics[width=1in,height=1.25in,clip,keepaspectratio]{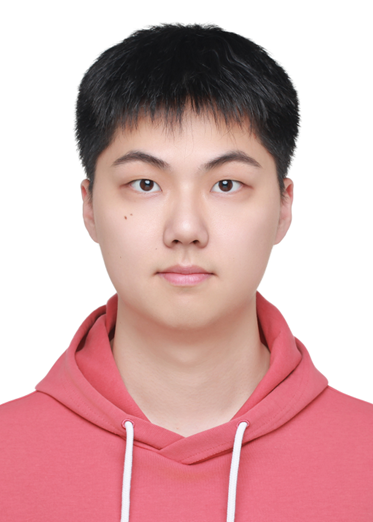}}]{Ziyao Li}
received his B.S. degree in communication engineering from East China Normal University in Shanghai, China in 2022, and and the M.S. degree in vehicle engineering with the School of Automotive Studies, Tongji University, Shanghai, China in 2025.

His research interests include Global Navigation Satellite System (GNSS) positioning and deep learning.
\end{IEEEbiography}

\vspace{-20pt}
\begin{IEEEbiography}[{\includegraphics[width=1in,height=1.25in,clip,keepaspectratio]{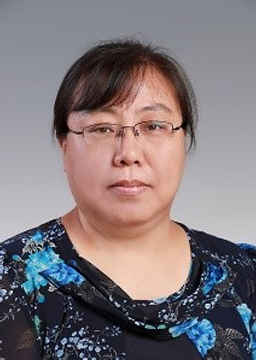}}]{Guirong Zhuo}
received her B.S. degree in Yanshan University in Hebei, China in 1990, and the M.S. and Ph.D. degrees in Harbin Institute of Technology in Harbin, China, in 1999 and 2002, respectively.

She is an associate professor and doctoral supervisor with the School of Automotive Engineering, Tongji University. Her research interests include vehicle dynamics control and intelligent vehicle combination positioning.
\end{IEEEbiography}

\vspace{-20pt}
\begin{IEEEbiography}[{\includegraphics[width=1in,height=1.25in,clip,keepaspectratio]{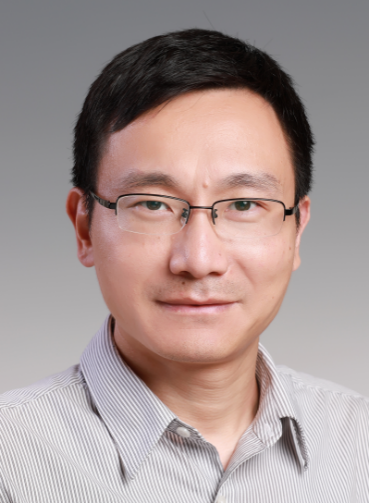}}]{Lu Xiong}
received the B.E., M.E., and Ph.D. degrees in vehicle engineering from School of Automotive Studies, Tongji University, Shanghai, China, in 1999, 2002, and 2005, respectively. 

He is currently a Professor with Tongji University, where he is also an Director of the School of Automotive Studies. His research interests include perception, decision and planning, dynamics control and state estimation, and testing and evaluation of autonomous vehicles.
\end{IEEEbiography}

\end{document}